%% file: main.tex
\documentclass[10pt,twocolumn,letterpaper]{article}

\usepackage{cvpr}              

\usepackage{graphicx}
\usepackage{amsmath}
\usepackage{amssymb}
\usepackage{booktabs}
\usepackage[pagebackref,breaklinks,colorlinks]{hyperref}

\usepackage[capitalize]{cleveref}
\crefname{section}{sec.}{Secs.}
\Crefname{section}{Section}{Sections}
\Crefname{table}{Table}{Tables}
\crefname{table}{Tab.}{Tabs.}

\usepackage{booktabs}
\usepackage{multirow}
\usepackage{color}
\usepackage{eucal}
\usepackage{overpic}
\usepackage{bm}

\usepackage[dvipsnames]{xcolor}
\def\cvprPaperID{3576} 
\def\confName{CVPR}
\def\confYear{2022}

\begin{document}

\title{Few-Shot Head Swapping in the Wild}

\author{Changyong Shu$^{1}$ \quad  Hemao Wu$^{2}$ \quad Hang Zhou$^{1}$\thanks{Corresponding authors.} \quad Jiaming Liu$^{1}$\footnotemark[1]  \quad  Zhibin Hong$^{1}$\\
Changxing Ding$^{2}$ \quad   Junyu Han$^{1}$ \quad
Jingtuo Liu$^{1}$ \quad Errui Ding$^{1}$ \quad Jingdong Wang$^{1}$ \\
$^1$Department of Computer Vision Technology (VIS), Baidu Inc.,  $^2$South China University of Technology \\
     {\tt\small \{zhouhang09,liujiaming03,hongzhibin,hanjunyu,liujingtuo,dingerrui,wangjingdong\}@baidu.com}, \\
      {\tt\small changyong.shu89@gmail.com, \{201830252427@mail.,chxding@\}scut.edu.cn. }
\vspace{-20pt}
}

\makeatletter
\vspace{-24pt}
\let\@oldmaketitle\@maketitle
\renewcommand{\@maketitle}{\@oldmaketitle
\centering

    \begin{overpic}[width=0.88\textwidth]{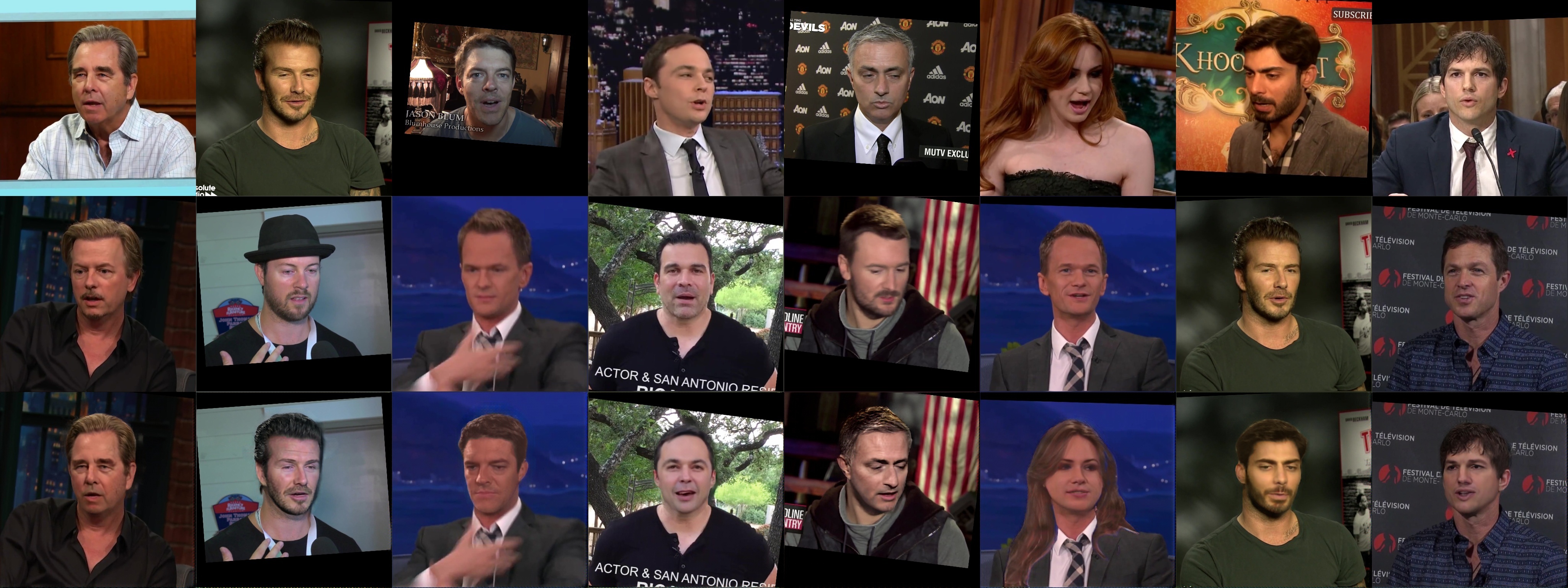}
        \put(-2.0, 27){\rotatebox{90}{\color{black}{\scriptsize Source Image}}}
        \put(-2.0, 14){\rotatebox{90}{\color{black}{\scriptsize Target Image}}}
        \put(-2.0,  1){\rotatebox{90}{\color{black}{\scriptsize Head Swapping}}}
    \end{overpic}

    \captionof{figure}{
    \textbf{Head swapping results generated by HeadSwapper}. The first line is the source image. The second line is the target image. The head swapping results are shown in the third line, where the source head is seamlessly transferred to the target in the wild.
    }
    \label{fig:fig1}
    \bigskip}                   
\makeatother
\maketitle


\begin{abstract}
The head swapping task aims at flawlessly placing a source head onto a target body, which is of great importance to various entertainment scenarios.
While face swapping has drawn much attention, the task of head swapping has rarely been explored, particularly under the few-shot setting. It is inherently challenging due to its unique needs in head modeling and background blending. 
In this paper, we present the \textbf{Head Swapper (HeSer)}, which achieves few-shot head swapping in the wild through two delicately designed modules. 
Firstly, a Head2Head Aligner is devised to holistically migrate pose and expression information from the target to the source head by examining multi-scale information. 
Secondly, to tackle the challenges of skin color variations and head-background mismatches in the swapping procedure, a Head2Scene Blender is introduced to simultaneously modify facial skin color and fill mismatched gaps in the background around the head. Particularly, seamless blending is achieved with the help of a Semantic-Guided Color Reference Creation procedure and a Blending UNet.
Extensive experiments demonstrate that the proposed method produces superior head swapping results in a variety of scenes\footnote{Demo videos and code are available at \url{https://jmliu88.github.io/HeSer}.}. 
\end{abstract}

\input{sections/intro}

\input{sections/related}

\input{sections/method}

\begin{figure*}[t]
\flushright
    \begin{overpic}[width=0.95\textwidth]{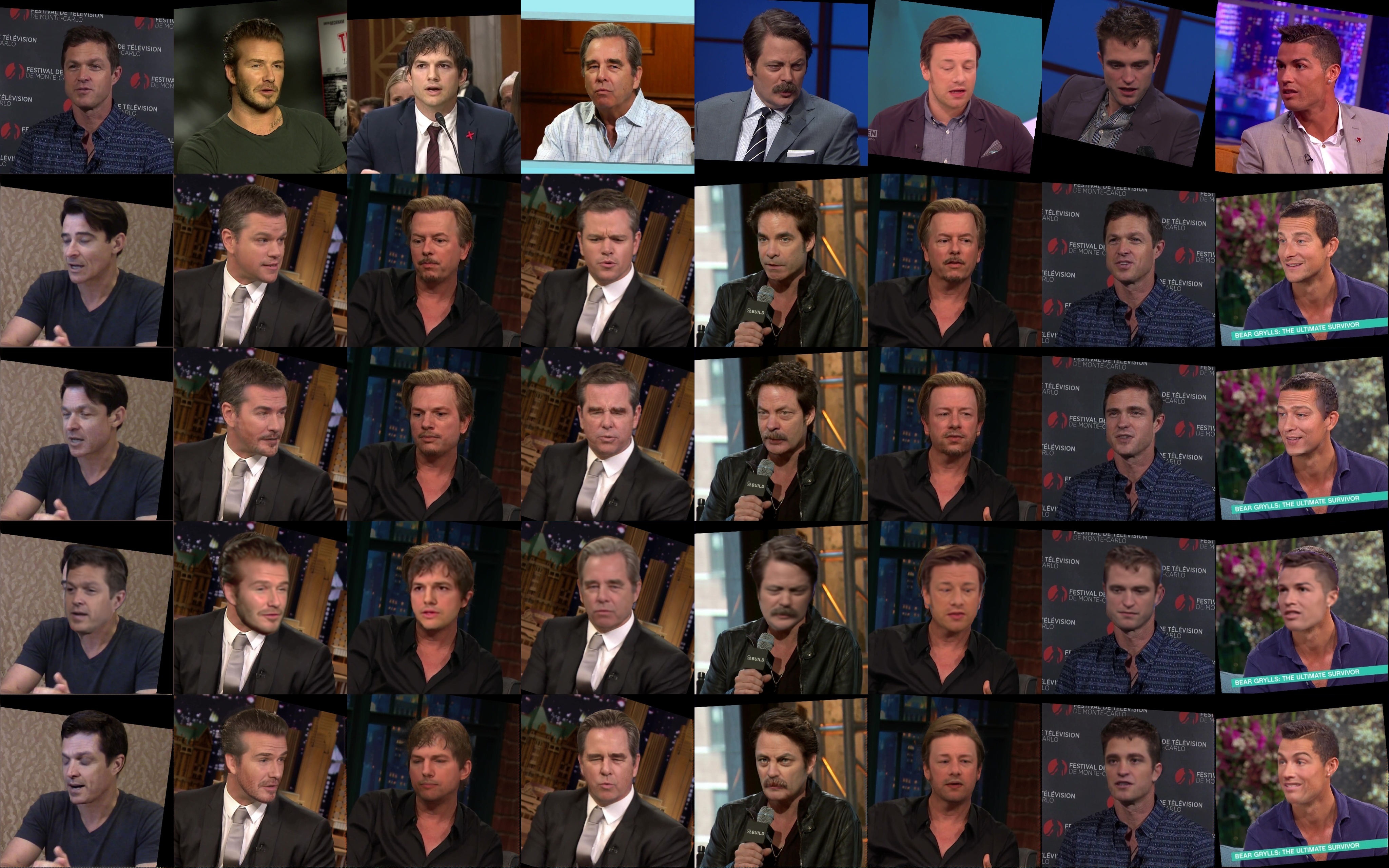}
        \put(-2,52.5){\rotatebox{90}{\color{black}{\scriptsize Source image}}}
        \put(-2,40.0){\rotatebox{90}{\color{black}{\scriptsize Target image}}}
        \put(-2,25.5){\rotatebox{90}{\color{black}{\scriptsize Face swapping \cite{chen2020simswap}}}}
        \put(-2,13.5){\rotatebox{90}{\color{black}{\scriptsize Deepfacelab \cite{Perov2020deepfacelab}}}}
        \put(-2, 4.5){\rotatebox{90}{\color{black}{\scriptsize Ours}}}
    \end{overpic}
    \vspace{-0.05in}
	\caption{Our head swapping results compared with other methods. 
	}
	\label{fig:compare_of_headswap_and_faceswap_and_deepfacelab}
\end{figure*}

\section{Experiments}

Detailed descriptions of the data collection, along with more experimental details, are elaborated in the supplementary material. 

\subsection{Headswap Results}
We first evaluate our head swapping result.
We compare our proposed HeSer with the state-of-the-art face swapping model \cite{chen2020simswap} and Deepfacelab \cite{Perov2020deepfacelab} as follows.

\noindent\textbf{Qualitative comparison.} 
The results in Fig. \ref{fig:compare_of_headswap_and_faceswap_and_deepfacelab} demonstrate that our method can
outperform other methods 
in multiple aspects, e.g., identity preservation, pose
and expression consistency, skin color alignment, head-background coherence, and fidelity. 
The source image for face swapping is one of the 32-shot images,
while the source images for Deepfacelab and our head swapping model are the same 32 frames.
The face swapping method can swap the facial elements from source image to the target image, and the mustache can be slightly transferred; however, the immutable face shape and hairstyle restrict the identity similarity in human cognition. 
Deepfacelab tends to animate poorly when there is no source image with a similar pose to target image; 
furthermore, it is incapable of inpainting the missing pixel and the performance of head-color alignment is significantly inferior to ours.
For more vividly head swapping results generated by our method please refer to the supplementary video.

\noindent\textbf{Quantitative Comparison.} 
We conduct a user study for quantitative comparison of our HeSer with the face swapping method \cite{chen2020simswap} and Deepfacelab \cite{Perov2020deepfacelab}. We ask the users to rank 1) how well the \textbf{ID} information is preserved in each method. 2) The emotion and pose similarity with the target image, we denote the results as \textbf{Exp}. 3) The consistency of the \textbf{Skin Color} between generated examples and the target torso. 4) The \textbf{Inpainting} smoothness between generated head and the background. 5) The \textbf{Holistic} quality of the generated frames.

We randomly select 25 groups of source images and 10 target images from the VoxCeleb2 test set, and totally 250 results are obtained for each method. 
Then 20 different workers, including 15 normal users and 5 experts, are asked to evaluate randomly selected 25 results for every dimension. The workers rank the methods according to the question, and the rankings are scaled into scores of zero to five, where worst is zero and best is five. 
Table \ref{tab:avg_score} shows the average score result from the user study. Our HeSer outperforms other methods by a large margin, except for the expression consistency and skin color alignment. Face swapping \cite{chen2020simswap} directly injects the identity information into the target image thus naturally keeping better facial structures and skin colors. 

\begin{table}[t] 
  \caption{Average score from the user study, rating from 0 to 5.}
  \label{tab:avg_score}
  \centering
   \renewcommand{\arraystretch}{1.2}
   \renewcommand{\tabcolsep}{3pt} 
  \resizebox{1.0\linewidth}{!}{
  \footnotesize
      \begin{tabular}{@{}lcccccc@{}}
        \toprule
         Method & ID $\uparrow$ & {Exp} $\uparrow$ & Skin Color $\uparrow$ & Inpainting $\uparrow$ & Holistic $\uparrow$  \\ \midrule
         FaceSwap~\cite{chen2020simswap}    & 1.22          & \textbf{3.02} & \textbf{3.36} & -- & 1.42 \\
         Deepfacelab~\cite{Perov2020deepfacelab} & 2.16          & 1.47          & 1.53 &  0.05      & 2.26 \\
         Ours                                    & \textbf{4.12} & 3.01         & 2.61 & \textbf{4.95} & \textbf{3.82} \\
        \bottomrule
      \end{tabular}
  }
\end{table}

\subsection{Superiority of the Head2Head Aligner}\label{Animated comparison}
In this subsection, we evaluate Head2Head Aligner independently.
As the Head2Head Aligner is devised to animate the source image with the pose and expression of the target image, which is similar to the task of reenactment, different reenactment strategies are compared.
The performance is discussed from two perspectives: K-shot inputs, and fine-tuning or not. As some of our 2D-based animated competitors (FOMM \cite{Siarohin2019fomm} and Siarohin et al. \cite{siarohin2021motion}) do not support few-shot inputs and fine-tuning, we split our evaluations into 1) a one-shot setting, where all competitors are compared; 2) otherwise, only LPD is compared.


\begin{table}[t] 
  \caption{Quantitative comparison on face reenactment.}
  \vspace{-0.05in}
  \label{tab:one-shot-meta-learing}
  \centering
   \renewcommand{\arraystretch}{1.2}
   \renewcommand{\tabcolsep}{3pt} 
  \resizebox{1.0\linewidth}{!}{
  \footnotesize
      \begin{tabular}{@{}lcccccc@{}}
        \toprule
         Method & $E_{ID}$ $\downarrow$ & $E_P$ $\downarrow$ & SSIM $\uparrow$ & LPIPS $\downarrow$ & PSNR $\uparrow$  \\ 
         \midrule
         FOMM~\cite{Siarohin2019fomm} & {0.49} & 0.275 & 0.76 & {0.18} & 30.92 \\
         LPD~\cite{Burkov2020lpd} & 0.71 & 0.063 & 0.52 & 0.50 & 28.84 \\
         Siarohin et al \cite{siarohin2021motion} & 0.71 & 0.137 & 0.73 & 0.20 & 30.01 \\
         Ours & 0.53 & {0.026} & {0.77} & 0.19 & {31.33} \\
         \hline
         LPD (32-shot + ft) & 0.23 & {0.024} & {0.62} & 0.36 & 29.46 \\
         Ours (32-shot + ft) & \textbf{0.22} & \textbf{0.024} & \textbf{0.89} & \textbf{0.12} & \textbf{33.26} \\
        \bottomrule
      \end{tabular}
  }
  \vspace{-0.05in}
\end{table}


\begin{figure}
\centering
    \begin{overpic}[width=0.90\linewidth]{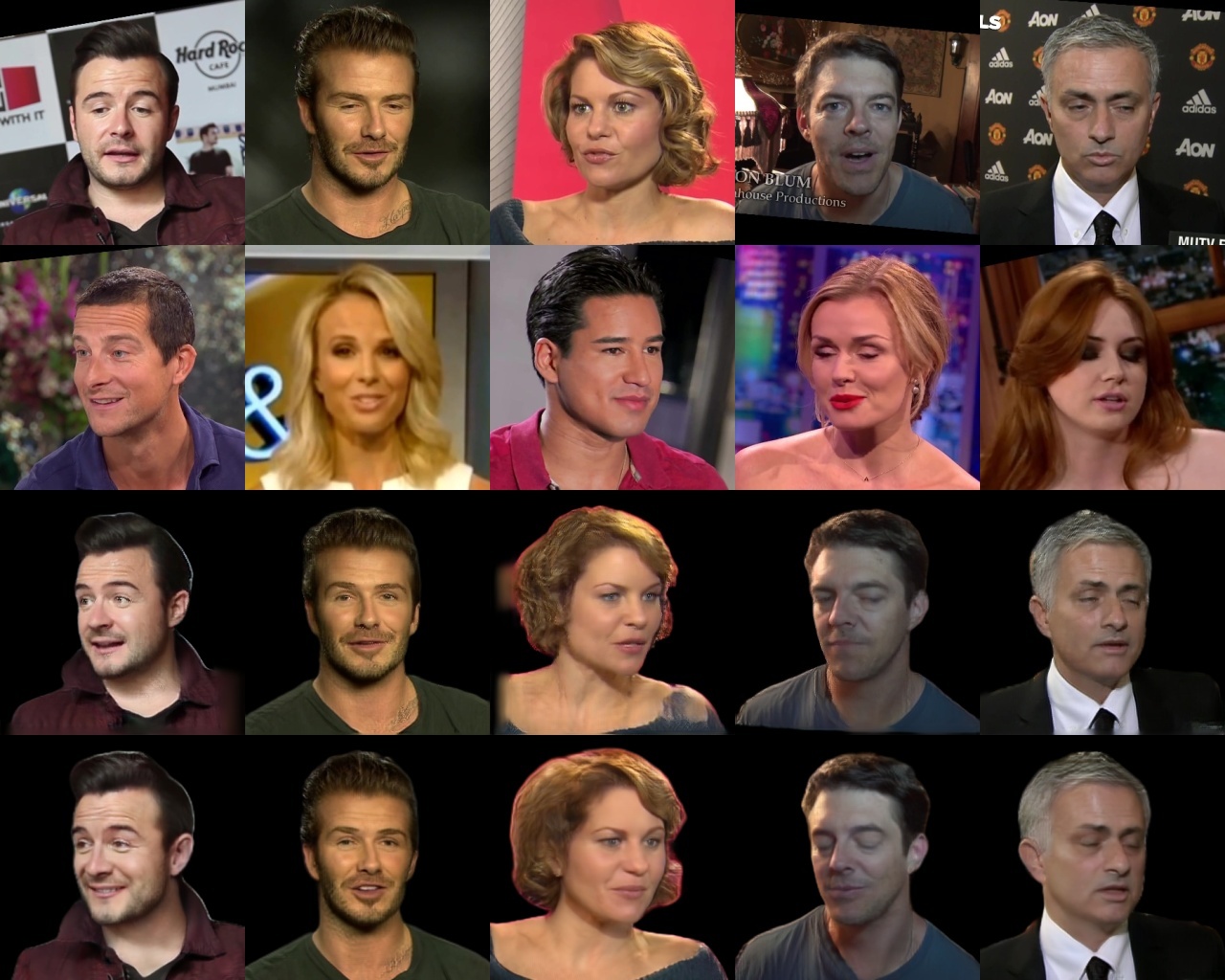}
        \put(36.0,81.0){\rotatebox{0}{\color{black}{\scriptsize 32-shot, ft600, cross-id}}}
        \put(-3.5,65.0){\rotatebox{90}{\color{black}{\scriptsize Source}}}
        \put(-3.5,47.0){\rotatebox{90}{\color{black}{\scriptsize Target}}}
        \put(-3.5,25.0){\rotatebox{90}{\color{black}{\scriptsize LPD \cite{Burkov2020lpd}}}}
        \put(-3.5, 7.0){\rotatebox{90}{\color{black}{\scriptsize Our}}}
    \end{overpic}
    \vspace{-0.05in}
	\caption{Qualitative cross-id animated results of 32-shot fine-tuned model with 600 iterations. 
	Source, target, and our animated result are re-aligned following the instruction of LPD.
	}
	\label{fig:32shot_ft600step_crossid}
\end{figure}


\begin{figure*}[t]
\centering
    \begin{overpic}[width=0.90\textwidth]{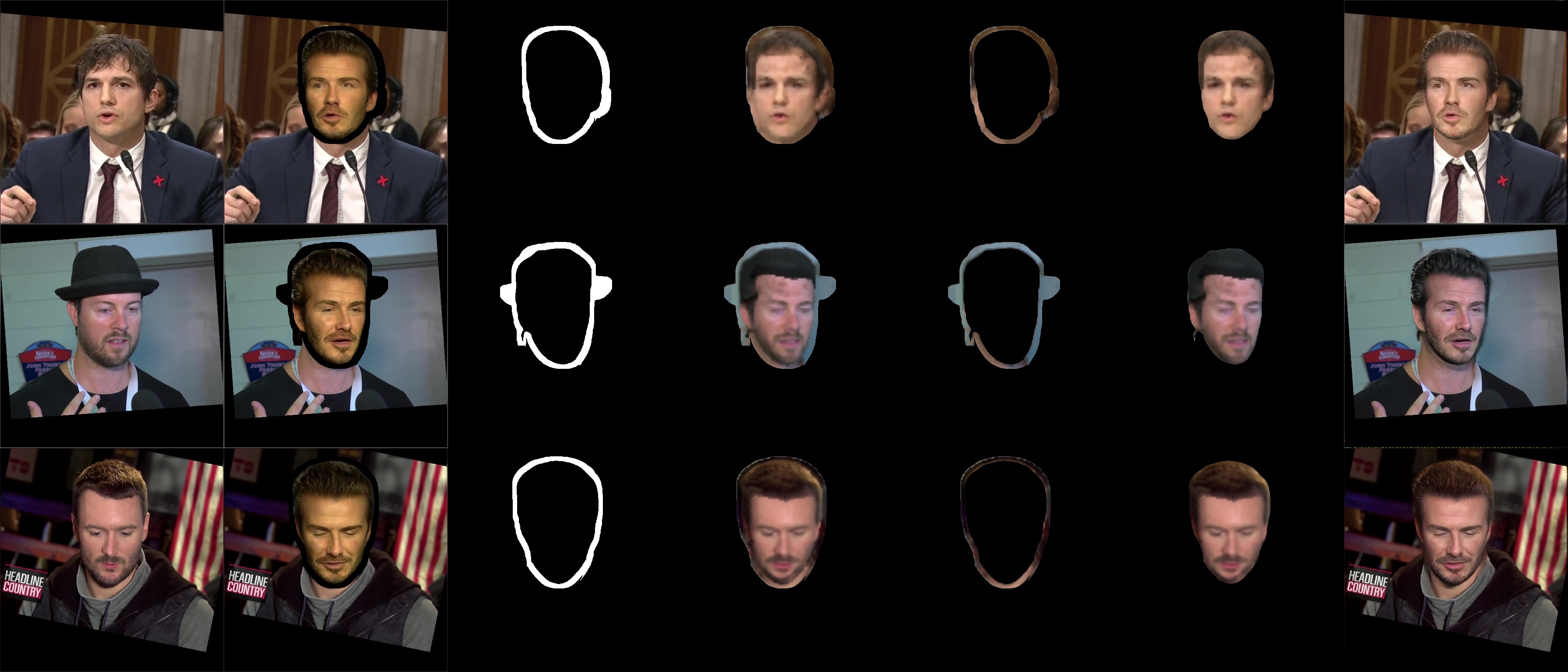}
        \put( 5.0,43.5){\rotatebox{0}{\color{black}{\scriptsize  Target}}}
        \put(15.0,43.5){\rotatebox{0}{\color{black}{\scriptsize  Animated portrait}}}
        \put(30.0,43.5){\rotatebox{0}{\color{black}{\scriptsize  Inpainting mask}}}
        \put(46.5,43.5){\rotatebox{0}{\color{black}{\scriptsize  Reference}}}
        \put(56.0,43.5){\rotatebox{0}{\color{black}{\scriptsize  Inpainting reference}}}
        \put(71.0,43.5){\rotatebox{0}{\color{black}{\scriptsize  Head-color reference}}}
        \put(87.2,43.5){\rotatebox{0}{\color{black}{\scriptsize  Blending result}}}
    \end{overpic}
	\caption{Illustration of head-color/inpainting reference in our Head2Scene Blender. Best viewed in color. %
	}
	\vspace{-0.05in}
	\label{fig:illustration_of_head-color_and_inpainting_exemplar}
\end{figure*}

\noindent\textbf{Evaluation Metrics.}
Five metrics are included in terms of head reenactment evaluation: (1) \textbf{Identity error} ($E_{ID}$) using the cosine similarity between face embeddings.
(2) \textbf{Pose reconstruction error} ($E_P$), measuring the pose error between the synthesized and ground truth images as performed in \cite{Burkov2020lpd,Zakharov20fast}. (3) Pixel-wise Reconstruction fidelity using \textbf{PSNR}. (4) Semantic perceptual similarity via the AlexNet-based \textbf{LPIPS} metric \cite{zhang2018unreasonable}. (5) The perceived quality via the structural similarity index measure (\textbf{SSIM}) \cite{zhou2004ssim}.

\noindent\textbf{One-shot setting Evaluation.}
We first compare the performance of our animated method with four competitors via five evaluation metrics under the one-shot face reenactment setting. The quantitative results are shown in Table. \ref{tab:one-shot-meta-learing}. Our method substantially outperforms all other animated models on every metric, except for the identity error and LPIPS, which is slightly lower than FOMM. For more qualitative comparison results please refer to supplementary material.


\noindent\textbf{Impact of K-shot and fine-tuning.} 
We further analyze the effect of increasing the K-shot number with subject-specific fine-tuning or a meta-learned model, which is the setting illustrated in Fig.~\ref{fig:fig1}. Specifically, we set the model that takes 32 frames as input and leverage 600 iterations of finetuning as the standard-setting. The quantitative comparisons are illustrated in Table~\ref{tab:one-shot-meta-learing}.
More quantitative evaluations on the influences of shot number and finetuning iterations are plotted in the supplementary material. It shows that our method almost outperforms other methods on all metrics with various settings. The performances of all metrics tend to improve as the k-shot number increases and fine-tuning is added respectively, while the identity error and pose reconstruction error of our animated method exhibit prominent improvement relative to LPD, demonstrating that our proposed scheme is capable of generating animated portraits with smaller identity gap and higher pose consistency. 

It is further noteworthy that our animated scheme outperforms LPD with only about one-fifth amounts of the training data in LPD. Though the identity error of our method and LPD is almost equal when the number of tuning steps is sufficiently large (such as 600), the facial expressions and emotions of our animated portraits are significantly better than those of LPD, as illustrated in Fig.~\ref{fig:32shot_ft600step_crossid}.
Please refer to supplementary material for more detailed qualitative results.

\subsection{Effectiveness of Head2Scene Blender}


The representative blending results are shown in the last column of Fig. \ref{fig:illustration_of_head-color_and_inpainting_exemplar}. Our Head2Scene Blender can produce head swapping portraits of photo-realistic quality. The head-color reference (sixth column in Fig. \ref{fig:illustration_of_head-color_and_inpainting_exemplar}) is created from the target image, while the head structure is consistent with the animated head. Similar behavior is observed in the inpainting reference (fifth column in Fig. \ref{fig:illustration_of_head-color_and_inpainting_exemplar}), where the estimated region style is faithful to the reference background. It is noteworthy that the hat region of the inpainting reference in the second row is constructed from the source background, which further demonstrates the inpainting effectiveness of our blending module.

\section{Conclusion and Discussion}
\noindent\textbf{Conclusion.} In this paper, we propose the Head Swapper (Heser), which achieves few-shot head swapping for in-the-wild scenarios. Specifically, our Head2Head Aligner generates high-fidelity reenact results with high pose and expression consistency, and the Head2Scene Blender seamlessly blends the aligned source head to the target image while maintaining the color of the target person. Extensive experiments demonstrate that the HeSer can achieve superior head swapping results on a variety of scenes.

\noindent\textbf{Broader Impact.}
Vivid video synthesis technologies create possibilities for immoral behaviors. Recent generative models have greatly influenced identity safe, image authenticity, etc. We will share the results of HeSer to the face/head forgery detection community for the healthy development of the AI technology.


\noindent\textbf{Acknowledgement.}
This work is supported by the CCF-Baidu Open Fund.

{\small
\bibliographystyle{ieee_fullname}
\bibliography{egbib}
}

\clearpage

\appendix

{\section*{Appendices}\huge }

\section{Head2Head Aligner Training Details}\label{Detailed_loss_functions_for_Head2Head_Aligner}

In the training stage of the Head2Head Aligner,  the target image $I_T$ and the source image set $I_S$ are all sampled from the same identity. We would expect the output animated portrait $I_A$ of our model to be the same as $I_T$, thus the whole training procedure is based on frame reconstruction. The loss functions are as follows:

\noindent\textbf{Pixel-wise Reconstruction Loss}. The reconstruction loss in the first stage $L^1_{L1}$ encourages the pixel-wise similarity between the reenactment output $I_R$ and the target image $I_T$ via L1 loss:
\begin{align}
    L^1_{L1} = \lambda_{L1}\left\|{I_T-I_A}\right\|_1.
\end{align}
where $\lambda_{rec}$ is the loss weight.

\noindent\textbf{Perceptual Loss}. 
We utilize the perceptual loss to minimize the semantic discrepancy between the animated portrait $I_A$ and the target image $I_T$. The feature matching loss is used on a pre-trained VGG-19 network, a pre-trained ResNet with ArcFace~\cite{deng2019arc} $E_{id}$, and the discriminator $D_A$:
\begin{equation}
\label{eq:perceptual}
    L^1_{per} = \sum^K_{k = 1}\sum^{L_k}_{l=1} \lambda^k_{l}\left\|{\phi^k_l(I_T)-\phi^k_l(I_A)}\right\|_1,
\end{equation}
where $\lambda^k_l$ balances the terms, $\phi^k_l$ represents the activation of layer $l$. $K = 3$ represents the three networks repectively.

\noindent\textbf{Identity Loss}. Moreover, to further enhance the ability of identity preservation, we leverage an additional identity loss to minimize the identity gap between the generated output and the target:
\begin{align}
    L_{id} = \lambda_{id}(1 - \cos{(E_{id}(I_T),E_{id}(I_A))}).
\end{align}
where $\lambda_{id}$ is the loss weight, $\cos$ denotes the cosine distance of identity embeddings.

\noindent\textbf{Adversarial Loss}. The discriminator $D_A$ is imposed to make the animated portrait $I_A$ look indistinguishable from the target image $I_T$. The training losses are defined as follows:
\begin{align}
	L^{D_A}_{adv} &= -E[h(D_A(I_T))] - E[h(-D_A(I_A))].
	\\
	L^{G_A}_{adv} &= -E[D_A(I_A)], 
    \label{eq:hinge_loss}
\end{align}
where $h(x)=\min(0,-1+x)$ is a hinge function used to regularize the discriminator~\cite{zhang2018self,brock2018large}. 

\section{Head2Scene Blender Training Details}\label{Supplementary_for_Head2Scene_Blender}

We jointly train the Semantic-Guided Color Reference Creation module and the Blending UNet via the loss functions below, expecting the sequential two jobs to facilitate each other. 

\noindent\textbf{Perceptual Loss.} Reconstruct training is also leveraged to minimize the difference
between the ultimate blending output ${I_{B}}$ and the target image ${I_T}$ by minimizing the perceptual loss.
\begin{equation}
    L^2_{per} = \sum^{L_1}_{l=1} \lambda^1_{l}\left\|{\phi^1_l(I_B)-\phi^1_l(I_T)}\right\|_1.
\end{equation}
here $\lambda^1_{l}$ balance the terms layer-wise, $\phi^1_l$ represents the activation of layer $l$ in the pre-trained VGG-19 as illustrated in Eq.~\ref{eq:perceptual}.

\noindent\textbf{Reconstruction Loss.} Note that the above feature matching loss excels in capturing fine details while missing the low-frequency image content. This could result in inaccurate colors. Consequently, the L1 loss is also applied for color consistency:
\begin{equation}
    L^2_{L1} = \lambda_{1}\left\|{I_B-I_T}\right\|_1.
\end{equation}

\noindent\textbf{Cycle Loss}: 
In order to guarantee that the warped headcolor/inpainting references could learn a meaningful correspondence matrix, we introduce the cycle consistent loss.
\begin{equation}
    L_c = \lambda_{c}\left\|{I_{T \rightarrow A \rightarrow T}-I_T}\right\|_1,
\end{equation}
where $I_{T \rightarrow A \rightarrow T}$ is the reference after cycled warpping, and $I_{T \rightarrow A \rightarrow T}^k(u) = \sum_{v \in M_A^k} {softmax}_v (\Gamma^k(u,v)/{\tau}) \cdot I_{T \rightarrow A}(v), u \in M_T^k$. Besides, the additional target image $I_T'$ coming from different image compared to $I_A$ is also utilized to ensure the meaningful of created reference:
\begin{equation}
    L_{c'} = \lambda_{c}\left\|{I_{T' \rightarrow A \rightarrow T'}-I_T}\right\|_1.
\end{equation}

\noindent\textbf{Adversarial Loss.} We utilize another discriminator $D_B$ to distinguish the blending outputs and the real samples from ground truth, with head mask $M_A^H$ and inpainting mask $M_A^I$ concatenated as conditions for further improving the fidelity of our blending outputs. The adversarial objectives are optimized by hinge loss:
\begin{align}
    \begin{split}
	L^{D_B}_{adv} &= -E[h(D_B(I_T \odot M_A^H \odot M_A^I))] \\
	&- E[h(-D_B(I_B \odot M_A^H \odot M_A^I))].
	\end{split}
	\\
	L^{B}_{adv} &= -E[D_B(I_B \odot M_A^H \odot M_A^I)].
    \label{eq:hinge_loss}
\end{align}
where $\odot$ denotes the concatenation along the dimensionality
of channel.

\begin{figure*}
\flushright
    \begin{overpic}[width=0.98\linewidth]{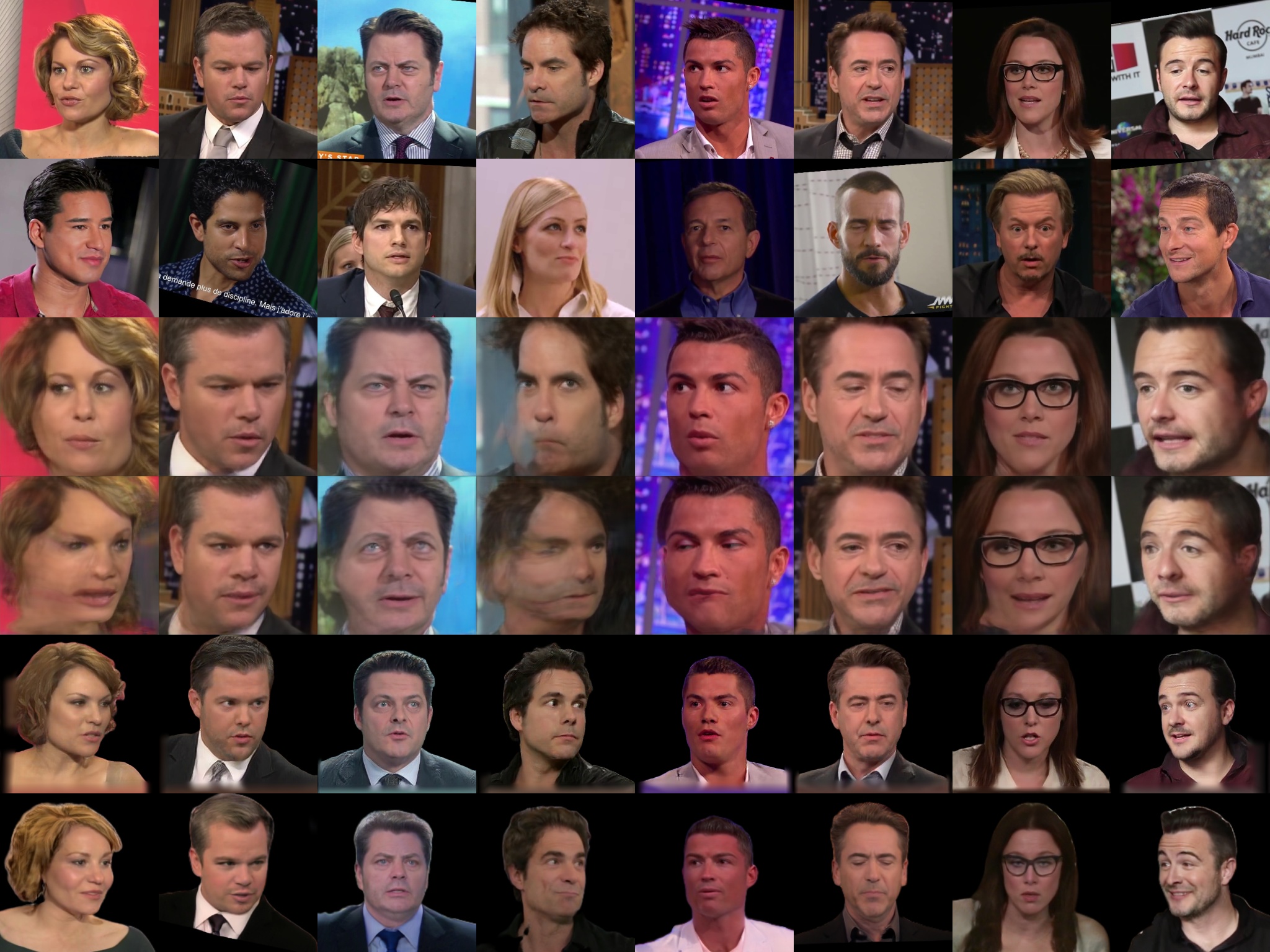}
        \put(45.5,75.5){\rotatebox{0}{\color{black}{\scriptsize 1-shot, meta, cross-id}}}
        \put(-1.5,66.0){\rotatebox{90}{\color{black}{\scriptsize Source}}}
        \put(-1.5,55.0){\rotatebox{90}{\color{black}{\scriptsize Target}}}
        \put(-1.5,41.0){\rotatebox{90}{\color{black}{\scriptsize FOMM}}}
        \put(-1.5,26.0){\rotatebox{90}{\color{black}{\scriptsize Siarohin et al \cite{siarohin2019animating}}}}
        \put(-1.5,17.0){\rotatebox{90}{\color{black}{\scriptsize LPD}}}
        \put(-1.5, 5.0){\rotatebox{90}{\color{black}{\scriptsize Our}}}
    \end{overpic}
	\caption{Qualitative cross-id animated portraits of one-shot meta-learned model. 
	1st row: source image for one-shot, 
	2nd row: target image from same video but different frame, 
	rows 3 through 6: animated result from FOMM, Siarohin et al \cite{siarohin2019animating}, LPD and ours.
	For a vivid show, our animated portraits are re-cropped following LPD.}
	\label{fig:1shot_meta_crossid_sup}
\end{figure*}

\begin{figure*}[!htb]
\centering
		\includegraphics[width=0.9\linewidth]{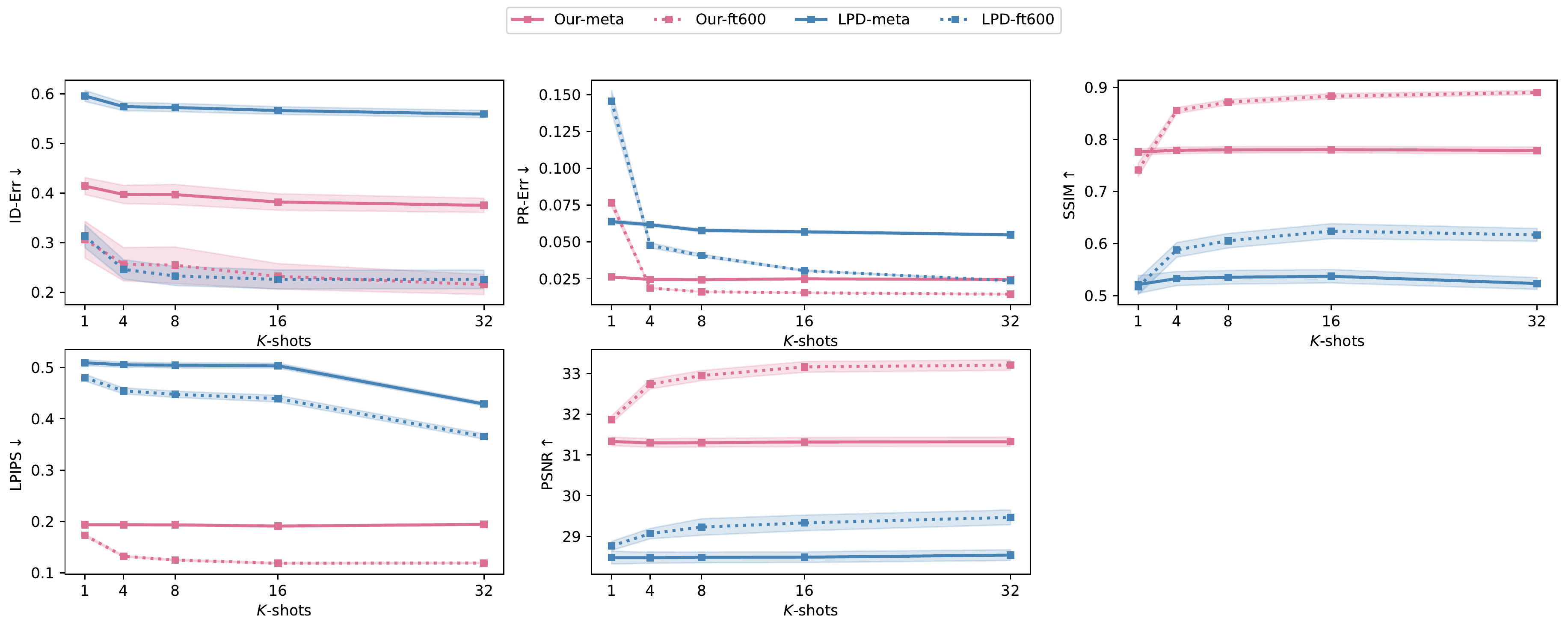}
	\caption{
	Quantitative result of increasing the K-shot number with subject-specific fine-tuning or a meta-learned model. Meta denotes the meta-learned model, ft600 indicates the fine-tuned model under 600 iterations. The transparent area around the dotted and solid lines represents the variance.
	}
	\label{fig:impact_of_kshot_and_ft}
	\vspace{-5pt}
\end{figure*}

\section{Experimental Details}\label{Datasets_and_implementation}

In this section, we describe the data collection and the experimental details, $i.e.$, evaluation metrics, competitors and implementation details. 

\subsection{Data collection}\label{data-collection}
In terms of Head2Head Aligner, 
We re-download the 1080P videos with the urls provided by the VoxCeleb2 dataset \cite{chung2018vox2}   from YouTube. 
The frames are aligned by detected landmarks and processed to ${512 \times 512}$.
Compared with previous studies~\cite{Burkov2020lpd,Zakharov19fsth}, a larger cropping window size is used.
Under such a setting, we manage to collect overall 28,367 videos with 5,478 different identities, 
which is much less than the original training set of 145,569 videos with 5,994 different identities in \cite{Burkov2020lpd,Zakharov19fsth}. 
Besides, totally 805 videos with 86 different identities are gathered as the test set. %

\subsection{Implementation Details.} 
The portrait encoder $E_{por}$ is ResNeXt-50~\cite{Xie17}, the pose and expression encoder are both constructed by the MobileNetV2 \cite{Sandler18}. The size of the latent embedding, $i.e.$ $d_1$, $d_2$, $d_3$ and $d_4$, are 512, 512, 256 and 256 respectively. The MLP module that transforms them into AdaIN parameters is a 2-layer ReLU perceptron with spectral normalization, where the output of the intermediate layer is 768. The generator is borrowed from \cite{Burkov2020lpd}. We add upsampling residual blocks after the last layer to generate the reenacted image with $512 \times 512$ resolution. The first module is trained for roughly three weeks with batch size to be 6 on six 32G Tesla V100 GPUs.

As for the Head2Scene Blender, we use a VGG-19 for feature extraction.
The Blending UNet is a seven-layer deep residual U-type network. The training image size is 512 $\times$ 512 and two 48G RTX8000 GPUs is used to train the model for 5 days.

For both models, Adam \cite{kingma2014adam} optimizer is used with $\beta_1=0.9$ and $\beta_2=0.999$, the imbalanced learning rate for generator and discriminator is set to be 1e-4 and 4e-4 respectively. Spectral normalization is applied to all operators in the system for stabilizing the adversarial training.

\begin{figure*}
\flushright
    \begin{overpic}[width=0.98\linewidth]{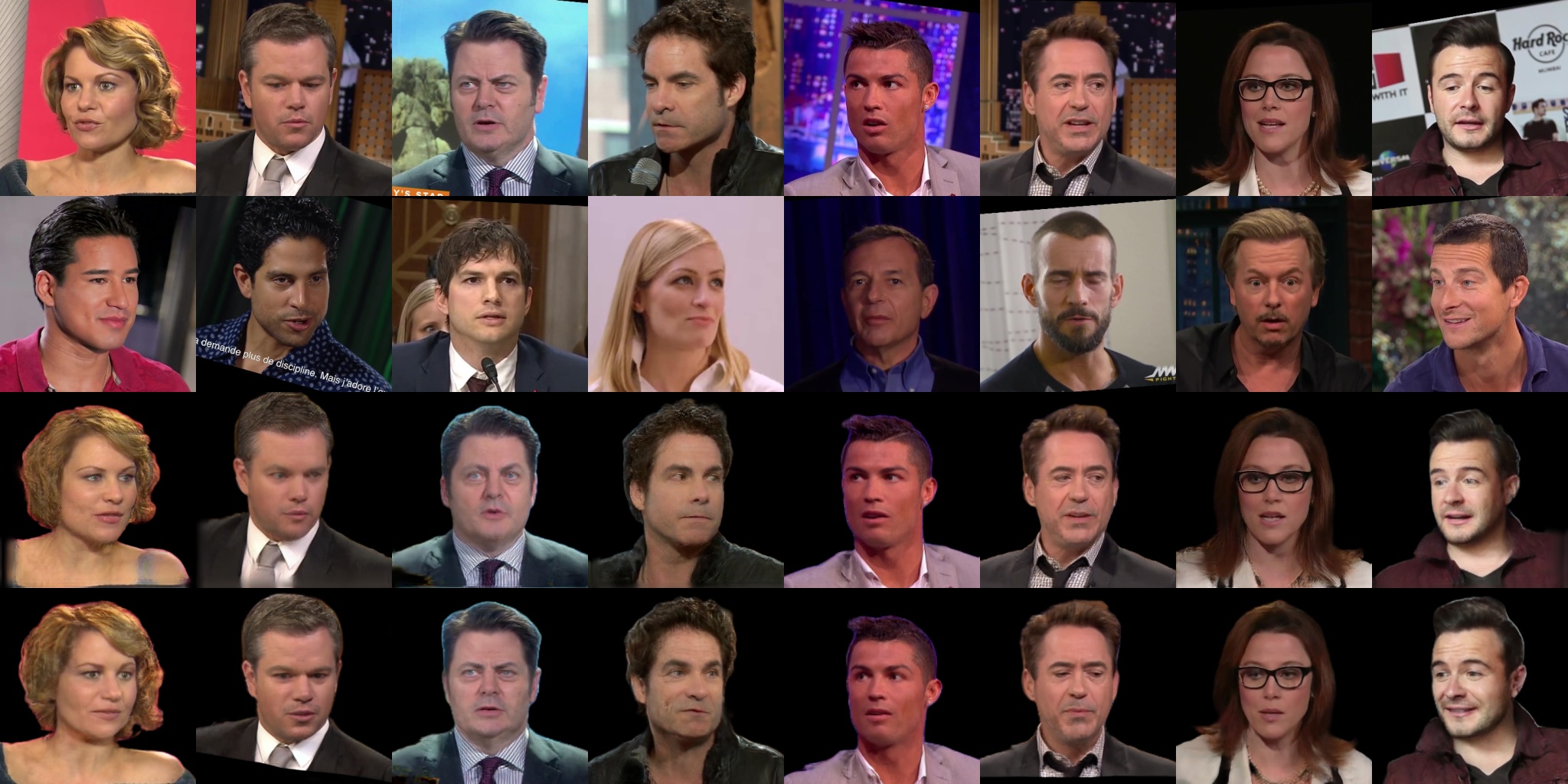}
        \put(45.5,50.5){\rotatebox{0}{\color{black}{\scriptsize 1-shot, ft600, cross-id}}}
        \put(-1.5,41.0){\rotatebox{90}{\color{black}{\scriptsize Source}}}
        \put(-1.5,30.0){\rotatebox{90}{\color{black}{\scriptsize Pose}}}
        \put(-1.5,17.0){\rotatebox{90}{\color{black}{\scriptsize LPD}}}
        \put(-1.5, 5.0){\rotatebox{90}{\color{black}{\scriptsize Our}}}
    \end{overpic}
	\caption{Qualitative cross-id animated portraits of one-shot finetune 600 iterations model, the layout is the same as in Fig. \ref{fig:1shot_meta_crossid_sup}.}
	\label{fig:1shot_ft600step_crossid_sup}
\end{figure*}

\begin{figure*}
\flushright
    \begin{overpic}[width=0.98\linewidth]{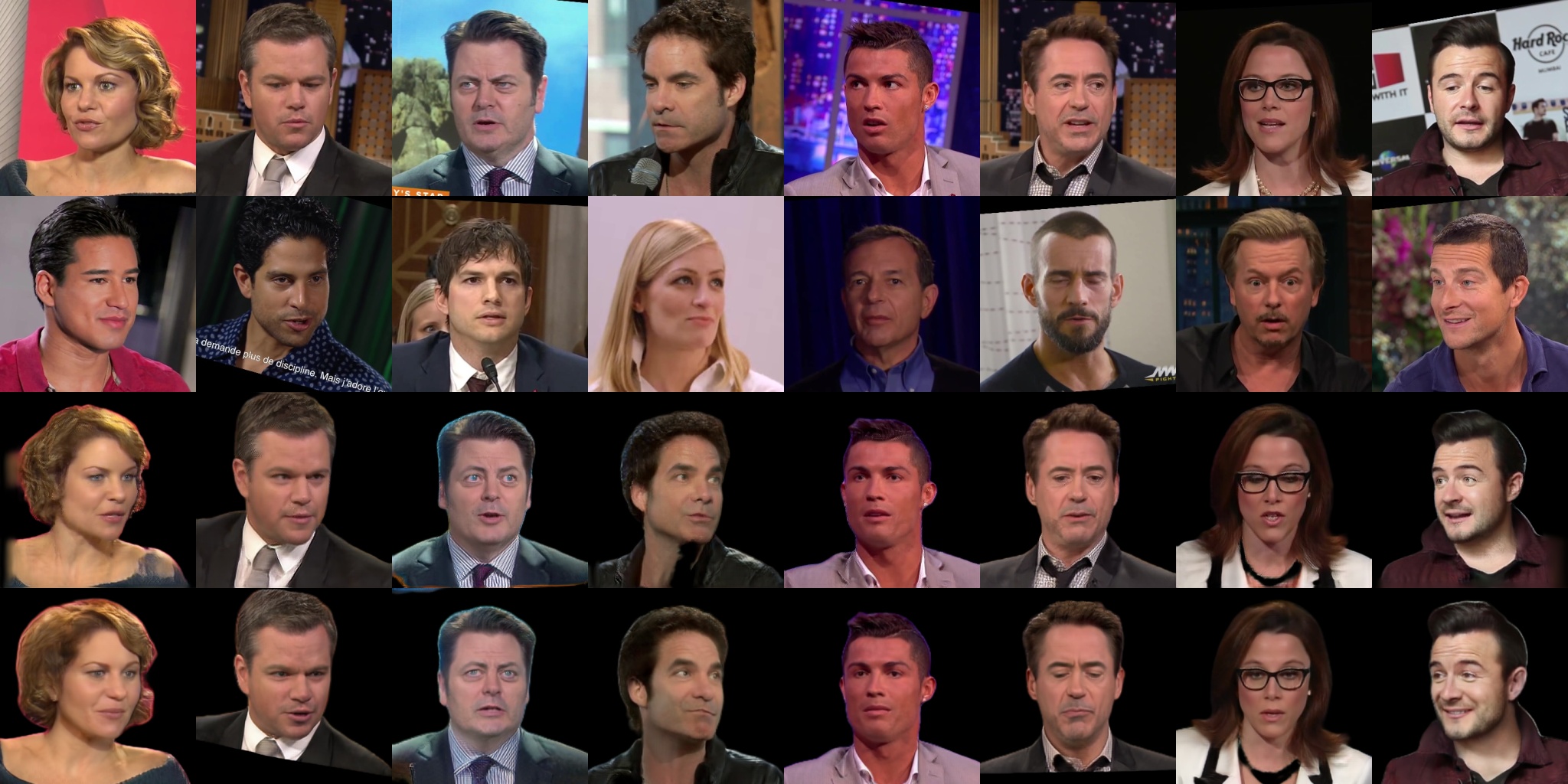}
        \put(45.5,50.5){\rotatebox{0}{\color{black}{\scriptsize 32-shot, ft600, cross-id}}}
        \put(-1.5,41.0){\rotatebox{90}{\color{black}{\scriptsize Source}}}
        \put(-1.5,30.0){\rotatebox{90}{\color{black}{\scriptsize Pose}}}
        \put(-1.5,17.0){\rotatebox{90}{\color{black}{\scriptsize LPD}}}
        \put(-1.5, 5.0){\rotatebox{90}{\color{black}{\scriptsize Our}}}
    \end{overpic}
	\caption{Qualitative cross-id animated portraits of 32-shot finetune 600 iterations model, the layout is the same as in Fig. \ref{fig:1shot_meta_crossid_sup}.}
	\label{fig:32shot_ft600step_crossid_sup}
\end{figure*}

\section{More Studies on Head2Head Aligner}\label{Supplementary_study_for_the_head2head_aligner}
This section is the supplementary materials for more quantitative and qualitative reenactment comparison under different settings:

\noindent\textbf{One-shot meta-learning evaluation.}
Fig. \ref{fig:1shot_meta_crossid_sup} is the supplementary qualitative results for cross-id animated portraits in one-shot meta-learning setting. Significant artifacts exhibited in FOMM and Siarohin et al \cite{siarohin2021motion}; our method can generate the animated portraits with higher identity similarity and pose consistency, besides, the elaborated emotions (such as happiness in 1st, 4th, and 8th row of Fig. \ref{fig:1shot_meta_crossid_sup}) are also well animated.

\noindent\textbf{Impact of K-shot and fine-tuning.}
Quantitative results of increasing the K-shot number with subject-specific fine-tuning or a meta-learned model are shown in Fig. \ref{fig:impact_of_kshot_and_ft}. The qualitative impact of fine-tuning is illustrated in Fig. \ref{fig:1shot_ft600step_crossid_sup}, where better identity preservation is obtained compared to the animated portrait from the meta-learned model, while the pose error is increased for the fine-tuned model overfits the one-shot source. Then we increase the K-shot number, as depicted in Fig. \ref{fig:32shot_ft600step_crossid_sup}, and the identity similarity and pose consistency are further improved.

\section{More Studies
 on Head2Scene Blender}
\label{Supplementary_analysis_of_the_head2background_blender}

This section is the supplementary materials for the discussion of 1) the memory-saving of our Semantic-Guided Color Reference Creation module, 2) skin color alignment and 3) inpainting performance in the blending network.

\begin{figure*}[htb]
\centering
    \begin{overpic}[width=1.0\linewidth]{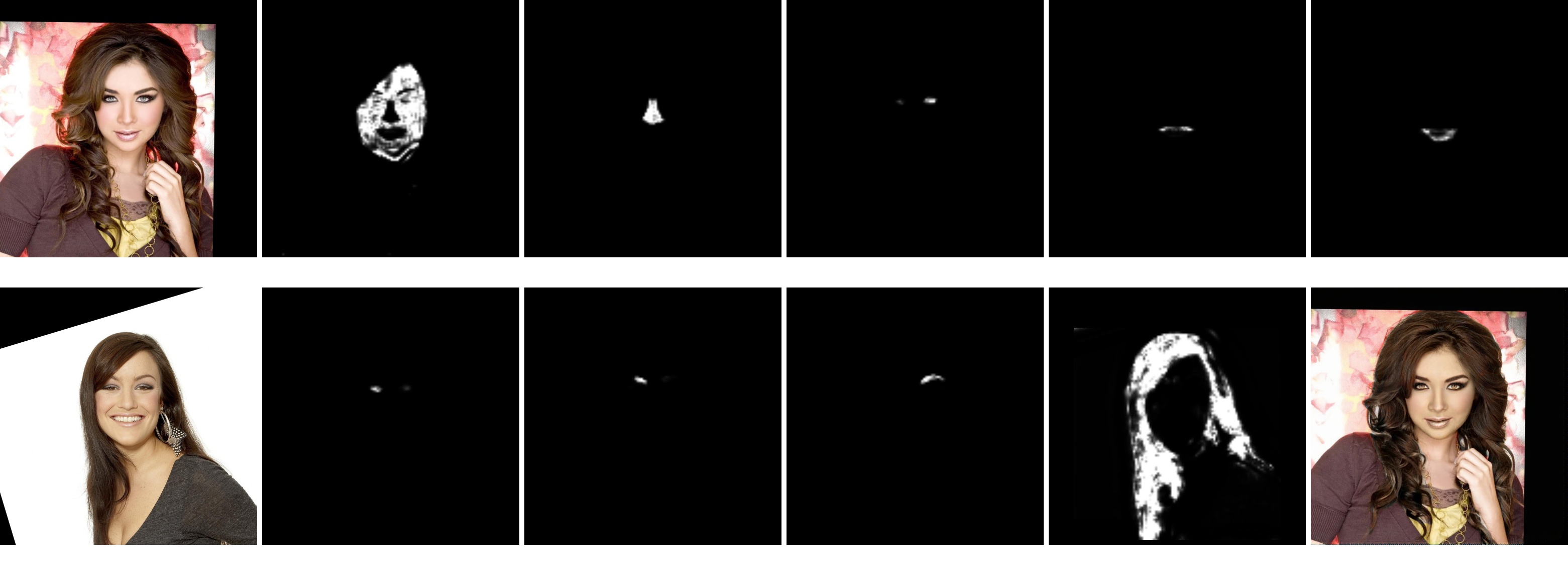}

        \put( 6.0,37.1){\rotatebox{0}{\color{black}{\scriptsize  source}}}
        \put(24.0,37.1){\rotatebox{0}{\color{black}{\scriptsize  skin}}}
        \put(40.0,37.1){\rotatebox{0}{\color{black}{\scriptsize  nose}}}
        \put(56.0,37.1){\rotatebox{0}{\color{black}{\scriptsize  right eye}}}
        \put(73.5,37.1){\rotatebox{0}{\color{black}{\scriptsize  up lip}}}
        \put(89.0,37.1){\rotatebox{0}{\color{black}{\scriptsize  low lip}}}

        \put( 5.0,18.6){\rotatebox{0}{\color{black}{\scriptsize  reference}}}
        \put(23.0,18.6){\rotatebox{0}{\color{black}{\scriptsize  left eye}}}
        \put(39.0,18.6){\rotatebox{0}{\color{black}{\scriptsize  left eye}}}
        \put(56.0,18.6){\rotatebox{0}{\color{black}{\scriptsize  left brow}}}
        \put(73.5,18.6){\rotatebox{0}{\color{black}{\scriptsize  hair}}}
        \put(90.0,18.6){\rotatebox{0}{\color{black}{\scriptsize  output}}}
        
    \end{overpic}
   \caption{
   The accumulated attention distribution for diverse semantic label, it shows an intuitive phenomenon: the highlighted regions correspond to semantic label blow the sub figure, and the remain regions are pitch-dark.
   }
   \label{fig:contribution}
\end{figure*}

\noindent\textbf{Semantic-Guided Color Reference Creation.} To verify the mechanism of our proposed calculation-reducing method, we present the semantic-specific contribution that the target image made in the correlation matrix. Specially, the accumulated attention distribution, denoting for pixels in source image with semantic label $k$, is computed via $\sum_{v \in M_T^k} {softmax}_v ( \Gamma^{k}(u,v) / \tau) (u \in M_A^k)$, and the qualitative results are exhibited in Fig.~\ref{fig:contribution}. Obviously, it demonstrates that the pixels in the target image with different semantic label almost has no contribution in the correspondence matrix to that in the source image. Thus the correspondence between pixels from source and target image with different semantic labels is redundant. Following our method, roughly averaged 9G GPU memory usage is saved in 512$\times$512 training setting.

\noindent\textbf{Skin Color Alignment.}
For effective comparison, we overlay the animated portrait on our head swapping result and using the target head as a reference. 
The qualitative results of skin color alignment with different method on VoxCeleb2 test set is illustrated in Fig. \ref{fig:makeuptransfer_voxceleb_half}, where the remarkable performance advantage is depicted.
Our method outperforms deepfacelab, SCGAN and PSGAN with more similar skin color to target image.

\begin{figure*}[htb]
\flushright
    \begin{overpic}[width=0.98\linewidth]{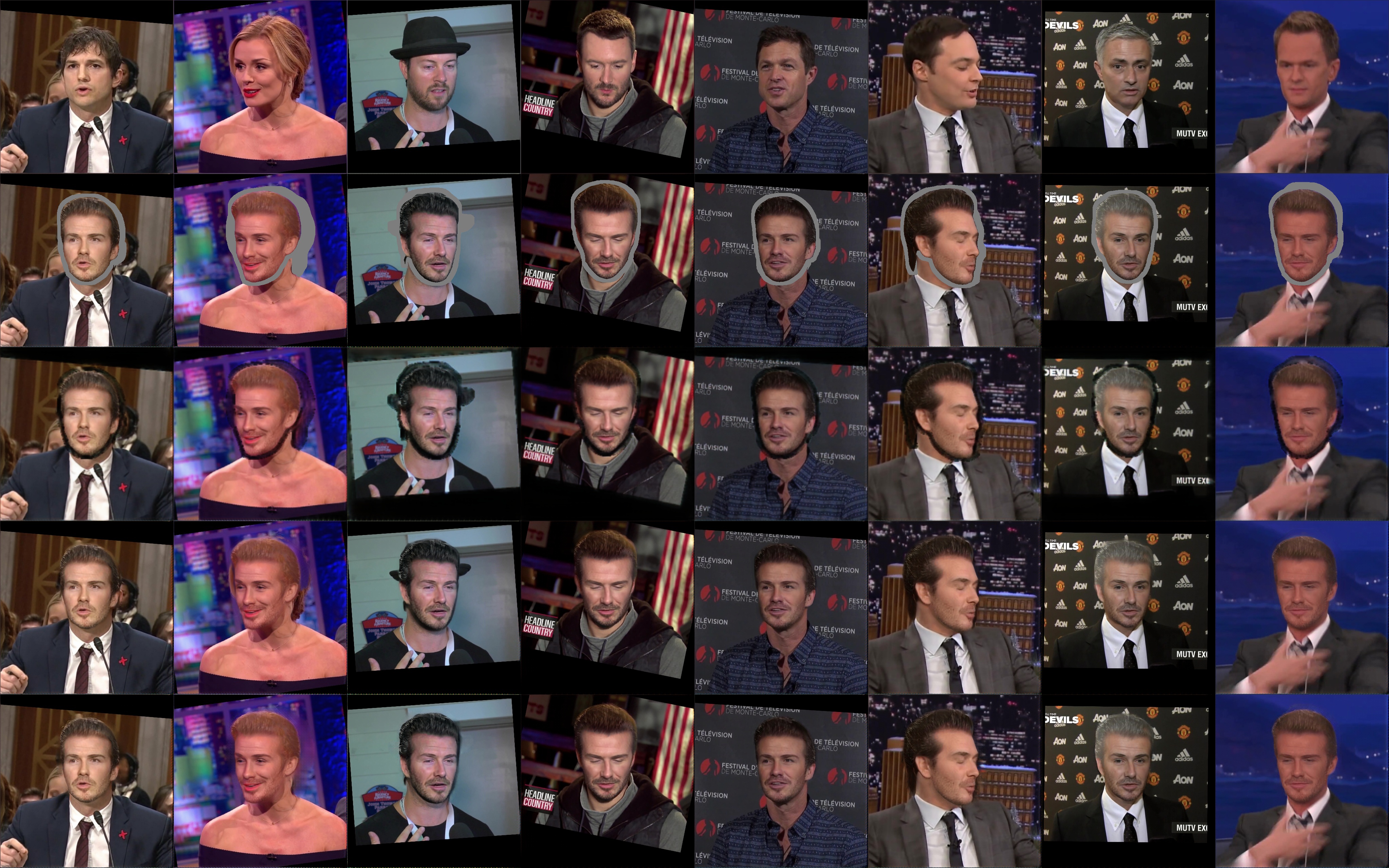}
        \put(-1.5,53){\rotatebox{90}{\color{black}{\scriptsize Reference}}}
        \put(-1.5,42){\rotatebox{90}{\color{black}{\scriptsize Source}}}
        \put(-1.5,26){\rotatebox{90}{\color{black}{\scriptsize Zhou et al. \cite{zhou2020loa}}}}
        \put(-1.5,14.5){\rotatebox{90}{\color{black}{\scriptsize TransFill \cite{zhou2021transfill}}}}
        \put(-1.5,4.5){\rotatebox{90}{\color{black}{\scriptsize Ours}}}
    \end{overpic}
   \caption{
   Reference-guided inpainting comparison.
   }
   \label{fig:paperfig_plot_for_inpaintingCompare}
\end{figure*}

\begin{figure*}[htb]
\centering
    \begin{overpic}[width=1.0\linewidth]{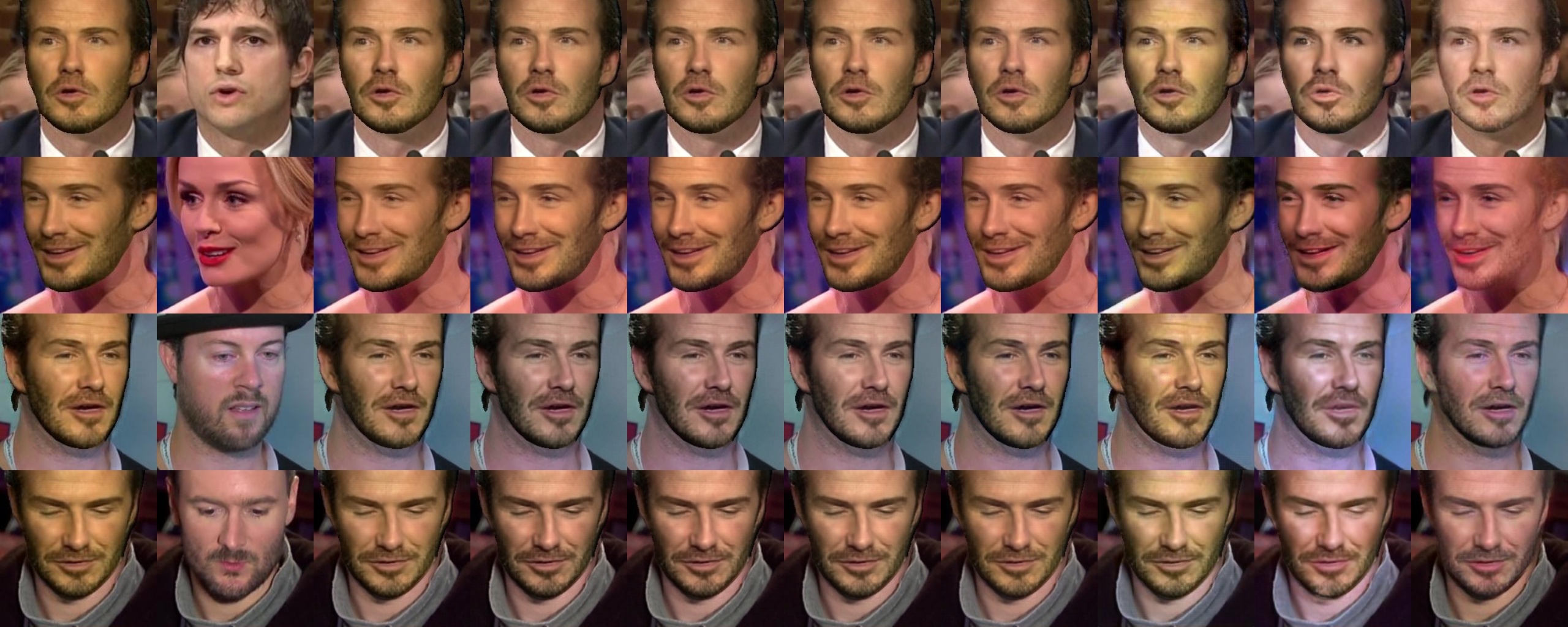}
        \put( 3.0,40.2){\rotatebox{0}{\color{black}{\scriptsize  Source}}}
        \put(12.0,40.2){\rotatebox{0}{\color{black}{\scriptsize  Reference}}}
    
        \put(22.0,40.2){\rotatebox{0}{\color{black}{\scriptsize  IDT \cite{Perov2020deepfacelab}}}}
        \put(33.0,40.2){\rotatebox{0}{\color{black}{\scriptsize  LCT \cite{Perov2020deepfacelab}}}}
        \put(42.5,40.2){\rotatebox{0}{\color{black}{\scriptsize  MKL \cite{Perov2020deepfacelab}}}}
        \put(52.0,40.2){\rotatebox{0}{\color{black}{\scriptsize  RCT \cite{Perov2020deepfacelab}}}}
        \put(62.0,40.2){\rotatebox{0}{\color{black}{\scriptsize  SOT \cite{Perov2020deepfacelab}}}}

        \put(71.0,40.2){\rotatebox{0}{\color{black}{\scriptsize  PSGAN\cite{jiang2020psgan}}}}
        \put(82.0,40.2){\rotatebox{0}{\color{black}{\scriptsize  SCGAN\cite{deng2021scgan}}}}
        \put(93.0,40.2){\rotatebox{0}{\color{black}{\scriptsize  Ours}}}
    \end{overpic}
   \caption{
   Qualitative comparsion of facial skin color alignment with different method on voxceleb2 test dataset, best viewed in color.
   }
   \label{fig:makeuptransfer_voxceleb_half}
\end{figure*}

\noindent\textbf{Performance of Inpainting.}
As the missing pixels are affected by the surroundings, for a fair comparison, we excavate neck and background around the head from our head swapping results and take them as input for inpainting, making facial skin color consistent with the skin color. 
Since the source code of reference-guided competitors ( LOA\cite{zhou2020loa} and TransFill\cite{zhou2021transfill}) are not released publicly, we send the testing data to authors for the compared result respectively. Qualitative comparison results are shown in Fig.\ref{fig:paperfig_plot_for_inpaintingCompare}.

\section{Limitations}

We briefly discuss the limitations of our two modules.
1) Although the Head2Head Aligner in HeSer can achieve better identity preservation and pose consistency than the previous 2D-based reenactment methods, it still needs a subject-specific fine-tuning procedure to generate an emotionally and positionally aligned head. 2) Segmentation masks are required for every frame of the target video, thus might lead to temporal inconsistency and increasing computational complexity. Moreover, HeSer performs poorly on long hair cases, 
which might be caused by the inaccuracy of segmentation results on the end of the hair.

Though there are certain limitations of our work, it is still one of the earliest attempts at few-shot head swapping, we hope it can contribute to the community and boost future head swapping techniques for solving the limitations above.

\begin{figure*}
\flushright
    \begin{overpic}[width=1.0\textwidth]{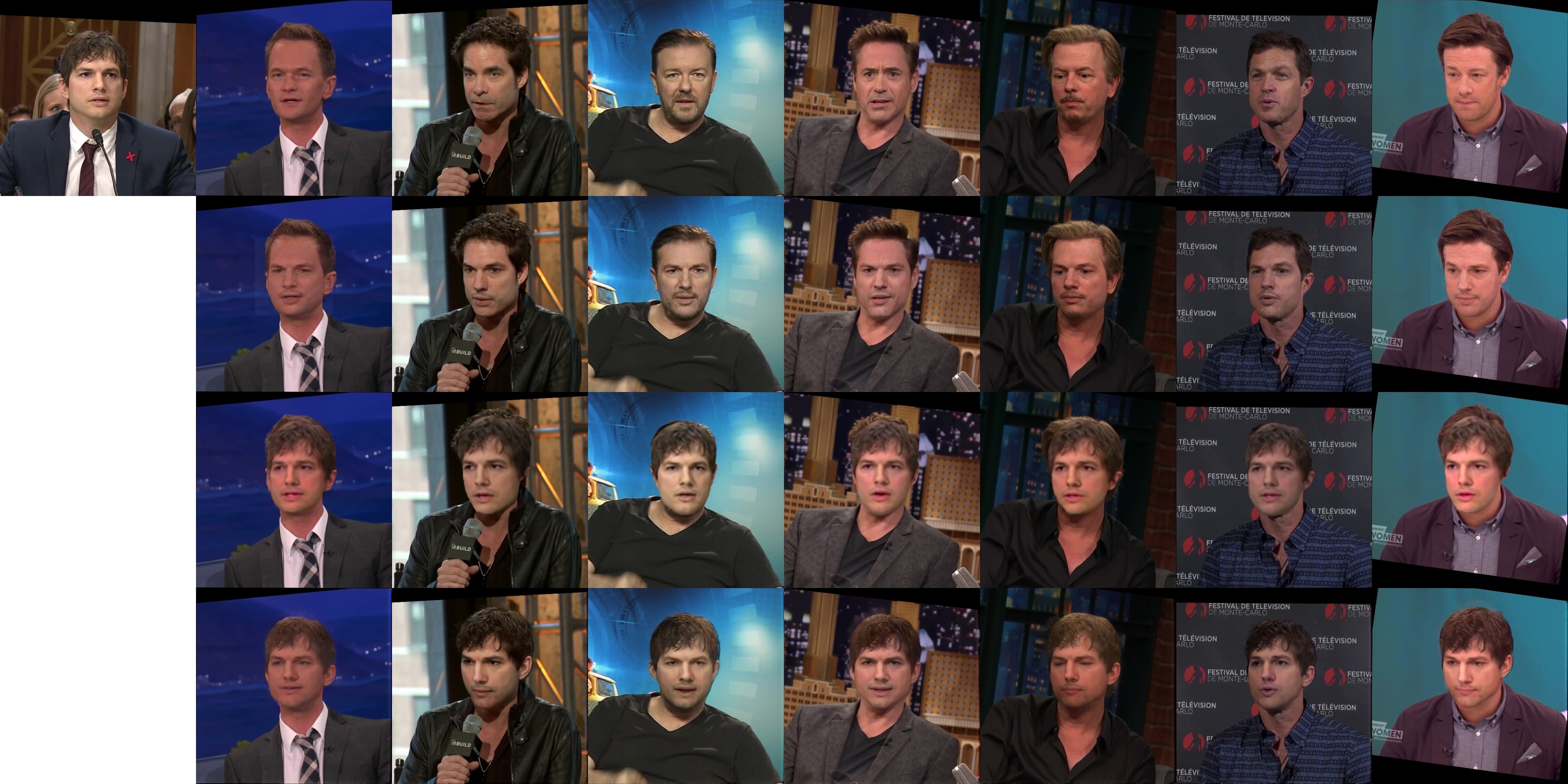}
        \put(0.5,30.5){\rotatebox{0}{\color{black}{\scriptsize Face swapping \cite{chen2020simswap}}}}
        \put(1.0,19.0){\rotatebox{0}{\color{black}{\scriptsize Deepfacelab \cite{Perov2020deepfacelab}}}}
        \put(4.0, 6.0){\rotatebox{0}{\color{black}{\scriptsize Ours}}}
    \end{overpic}
	\caption{Additional comparison results with other methods on VoxCeleb2 test set.
	}
	\label{fig:supp_headswap_7}
\end{figure*}
\end{document}

%% file: sections/intro.tex
\section{Introduction}\label{sec1}
Human cognition of identity appearance is profoundly affected
by not only
facial structures but also
head shapes and hairstyles.
Head swapping, the ability to seamlessly replace the head in a target image with a source one (as shown in Fig. \ref{fig:fig1}) would be of great importance to a variety of scenarios such as movie and advertisement composition, virtual humans creation, and deepfake video detection, etc.

While face swapping has long been a topic of interest~\cite{chen2020simswap,li2019fs,Nirkin2019fsgan,Perov2020deepfacelab,Zhu_2021_CVPR}, only a few studies have been carried out on the task of head swapping. 
DeepFaceLab \cite{Perov2020deepfacelab} requires large manual intervention to generate head swapping results, and 
StylePoseGAN \cite{sarkar2021styleposegan} tends to change the color of body skin and background in the target image in an undesired manner. 
Moreover, both methods fail to address few-shot head swapping, particularly for in-the-wild scenes.
%
%


We identify 
several properties that make few-shot head swapping more challenging than face swapping: 1) Head swapping requires not only perfect facial identity and expression modeling, but also capturing the structural information of a whole head and the non-rigid hair. 
Thus previous identity extraction strategies for face swapping~\cite{Nirkin2019fsgan,chen2020simswap} cannot be directly applied to head swapping. 2) There would be a huge \emph{region mismatch between swapped head edges and backgrounds} caused by the editing of head shapes and hairstyles. Such a problem does not exist in the face swapping setting. 3) Moreover, similar to face swapping, the color difference between source and target skins needs to be handled carefully.
%
%


In this paper, we propose a
framework called \textbf{Head Swapper (HeSer)}, which generates high fidelity head swapping results in the wild based on a few frames. Our key insight is to \emph{positionally and emotionally align the source head with the target in a unified blender that seamlessly handles both color and background mismatches.
}
Two modules, namely the \emph{Head2Head Aligner} and the \emph{Head2Scene Blender} are devised.
%
The Head2Head Aligner is responsible for finding a latent representation of a whole head as well as the facial details. It aligns the source head to the same pose and expression as the target image in a holistic manner. The identity, expression, and pose information are prominently balanced in a style-based generator by encoding multi-scale local and global information from both images. Moreover, a 
subject-specific fine-tuning
procedure could further improve identity preservation and pose consistency.

To further blend the aligned head into the target scene,
we devise a module named Head2Scene Blender, which provides both 1) color guidance for facial skins and 2) padding priors for  inpainting the gaps on the background around the head. Thus both the skin color and edge-backgrounds mismatches between source and target can be handled within one unified sub-module. It efficiently creates colored references by building the correlations between pixels of the same semantic regions.
Then with a blending UNet, seamless and realistic head swapping results can be rendered.
%


We summarize our main contributions as follows:
    \textbf{1)} We introduce a Head2Head Aligner that holistically migrates position and expression information from the target to the source head by examining multi-scale information.
    \textbf{2)} We design a
    Head2Scene Blender to simultaneously handle facial skin color and background texture mismatches.
    \textbf{3)} Our proposed \textbf{Head Swapper (HeSer)} produces photo-realistic head swapping results on different scenes. To the best of our knowledge, this is one of the earliest methods to achieve few-shot head swapping in the wild.

%% file: sections/related.tex
\section{Related Work}
The proposed two-stage Head Swapper is closely related to the topics of face swapping, head reenactment and image blending, which will be discussed below.

\noindent\textbf{Face and Head Swapping.}\label{rel-faceswapping-and-reenactment}
Many methods have been proposed for face swapping~\cite{nirkin2018face,Nirkin2019fsgan,Bao2018ipgan,Li2019faceshifter,chen2020simswap,Zhu_2021_CVPR,xu2021facecontroller,xu2022mobilefaceswap}. While few studies tackle the task in a source-oriented manner~\cite{nirkin2018face,Nirkin2019fsgan} that aims at reenacting and blending the source into the target, most methods are performed in a target-oriented manner \cite{Bao2018ipgan,Li2019faceshifter,chen2020simswap,xu2021facecontroller,Zhu_2021_CVPR}. They extract identity representation from the source image and inject it into the target image in a generative model.
Face swapping approaches produce immutable hairstyles and head shapes, which limits the overall similarity between the generated results and the source. 

The task of head swapping has been rarely studied. Deepfacelab \cite{Perov2020deepfacelab} is the first work to tackle head swapping.
However, it 
suffers from the following issues:
1) huge amounts of source data are required; 2) color transfer can only be  poorly performed; and 3) the regions that require inpainting the fusion look unnatural. 
Later StylePoseGAN \cite{sarkar2021styleposegan}, which is specially designed for re-rendering using pose-/appearance-conditioned StyleGAN \cite{karras2020stylegan2}, potentially has the capability of  head swapping. However, it tends to perform poorly for in-the-wild scenarios.
Our proposed head swapping pipeline is the first work committed to achieving few-shot head swapping in the wild.

\noindent\textbf{Few-Shot Head Reenactment.}\label{rel_Head_Reenactment}
Head swapping requires placing the source head at the target head's position with the same pose and expression, which is similar to the task of talking head generation and head reenactment~\cite{kim2018dvp, thies19dnr,gafni20dnrf,siarohin2019animating,Siarohin2019fomm,Burkov2020lpd,Chen_2020_CVPR,zhang2020freenet,zhang2021flow,Zhou2021CVPR,siarohin2021motion,JiCVPR2021,lu2021live,liang2022expressive}. Here we will briefly discuss the visual-driven few-shot reenactment methods.

Recent few-shot methods can be roughly divided into warping-based and reconstruction-based methods. 
Warping-based approaches\cite{siarohin2019animating,Siarohin2019fomm,siarohin2021motion} deform the source images to imitate the motion of driving ones, they tend to work poorly for the case with a big range of motion. Reconstruction-based approaches mostly leverage generative adversarial networks (GANs)~\cite{goodfellow2014generative}. A great number of studies use intermediate representations such as landmarks~\cite{Zakharov19fsth,Chen_2020_CVPR} and 3D models~\cite{zhang2021flow}. However, the inaccuracy of the structural information might lead to error accumulation. Certain recent studies focuses on the extraction and disentanglement of pose and identity information in the latent space~\cite{zhou2019talking,Burkov2020lpd,Zhou2021CVPR,liang2022expressive}. Particularly, LPD~\cite{Burkov2020lpd} verifies that a style-based generator can explicitly handles such information in latent vectors.  Inspired by this type of work, we introduce the Head2Head Aligner to find the latent representation of the whole head as well as the facial details in a reconstruction-based learning pipeline.


\begin{figure*}
\centering
		\includegraphics[width=0.85\linewidth]{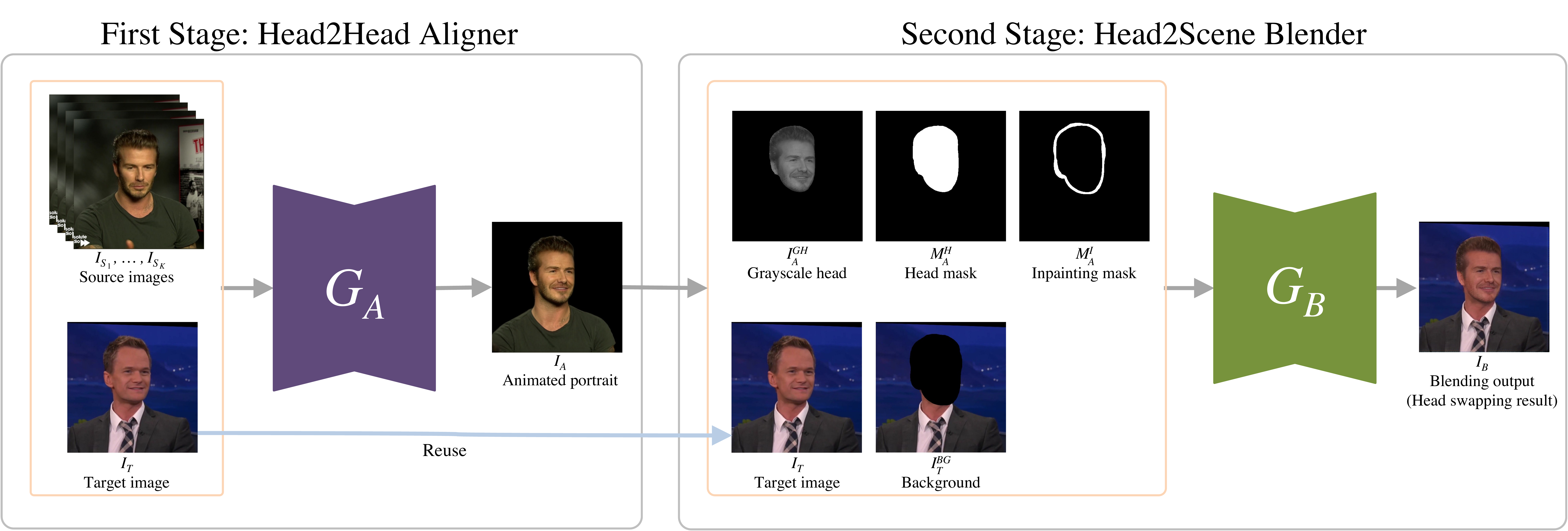}
	\caption{The overall architecture of our proposed pipeline for testing. First stage: the Head2Head Aligner $G_A$ (detailed in Fig. \ref{fig:lpd_framework_fortrain}) generates the animated portrait ${I_A}$, depicts the ${K}$-shot source image ${I_{{ID}_k}}(k=1,...,K)$ at the same pose and expression of the target  image ${I_T}$. 
	Second stage: we concatenate grayscale head $I_{GH}$, head mask ${M_H}$, inpainting mask $M_I$, reference source $I_T$ and background $I_{BG}$ to go through the Head2Scene blender $G_B$ (detailed in Fig. \ref{fig:blending_framework-fortest}) to generate the blending output $I_B$ (i.e., the head swapping result) in photo-realistic quality.
	}
	\label{fig:overview}
\end{figure*}

\noindent\textbf{Color Transfer.} As we need to transfer the skin color of the source face to the target one, it is similar to the task of makeup transfer.
Early makeup transfer methods utilize facial landmarks or parsing results as prior information~\cite{Change2018pairedcyclegan,chang2018pairedcyclegan,chen2019beautyglow,gu2019ladn,li2018beautygan}, however, they cannot keep the facial expressions unchanged.
The subsequently PSGAN \cite{jiang2020psgan,liu2021psgan++} and SCGAN \cite{deng2021scgan} solve the pose misalignment problem above. However, they tend to change the background color after the color transfer, making them unsuitable for head swapping where the background of the target must not be changed. 


%% file: sections/method.tex
\section{Method}
In this work, we introduce the Head Swapper (HeSer), which is illustrated in Fig. \ref{fig:overview}. Let $I_{S} = \{I_{S_k}|k \in [1, K]\}$ denote the set of source images, and let ${I_T}$ be the image of the target person. 
We aim to create a new blending output (\textit{i.e.}, head swapped result) ${I_B}$, with the head of $S$ on the body and background of ${I_T}$.
Notably, the pose, expression, and skin color of ${I_B}$ should remain consistent with ${I_T}$, but the identity, head structure and hairstyle should be the same as $S$.

In this section, we first design a \textbf{Head2Head Aligner} (Sec.~\ref{arch-aligner}) to produce an \emph{animated portrait} image $I_A$ as an intermediate representation. Then $I_B$ is derived by blending $I_A$ with $I_T$ in the \textbf{Head2Scene Blender} (Sec.~\ref{arch-blending}).

\begin{figure}[t]
\centering
      \includegraphics[width=0.95\linewidth]{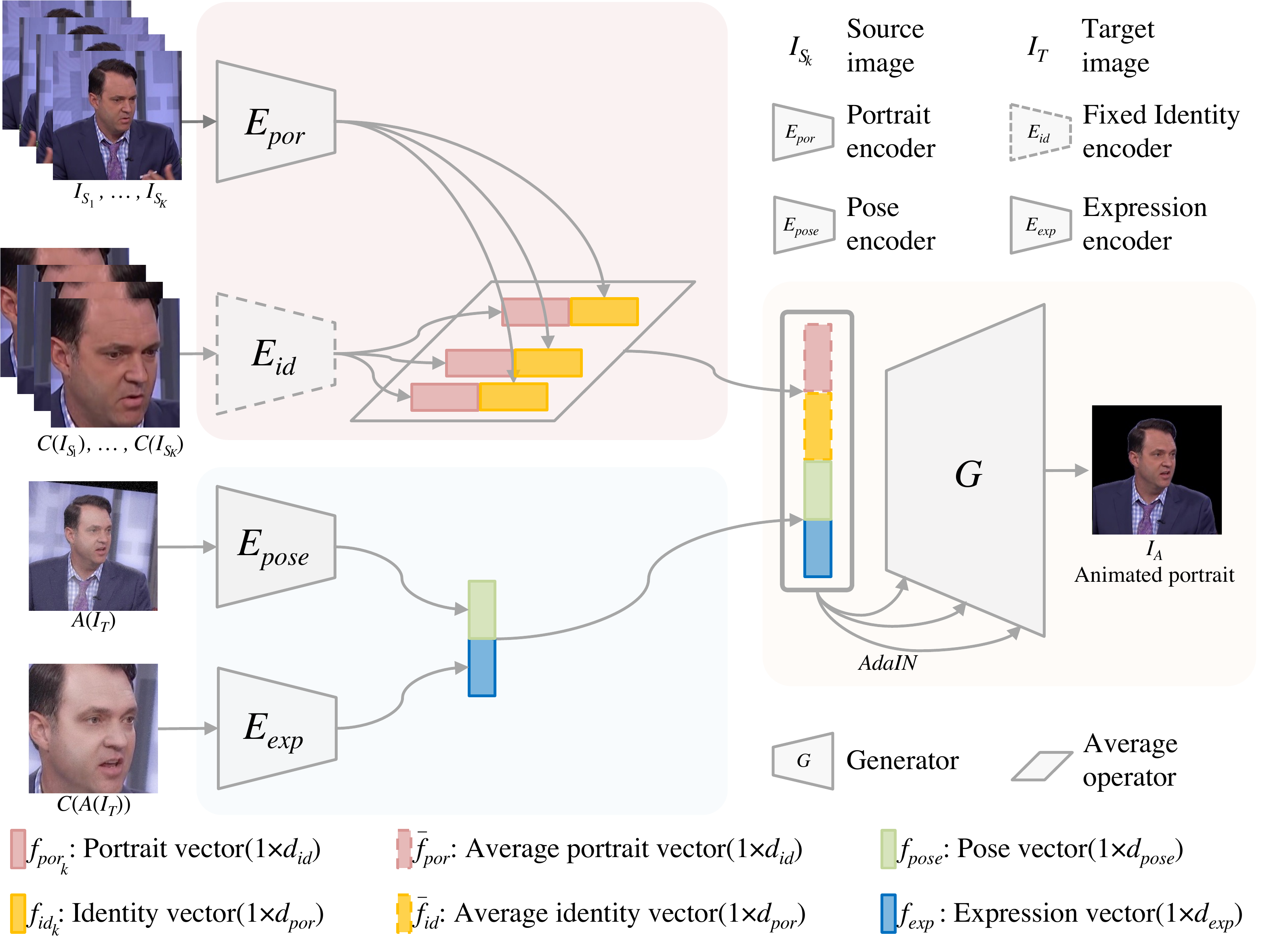}
   \caption{
   Overview of the proposed Head2Head Aligner for training. The average portrait vector ${{\it{\bar{f}_{por}}}}$, average identity vector ${{\it{\bar{f}}_{id}}}$, pose vector ${{\it{{f}}_{pose}}}$ and expression vector ${{\it{{f}}_{exp}}}$ are obtained from the portrait encoder, identity encoder, pose encoder and expression encoder respectively. Then the vectors above are sent into the animated portrait generation.
   }
   \label{fig:lpd_framework_fortrain}
   \vspace{-10pt}
\end{figure}

\subsection{Head2Head Aligner}\label{arch-aligner}

Instead of directly building $I_B$,
our first step is to produce an image which is positionally and emotionally aligned with $I_T$, but with the same identity as $S$. We name this intermediate \emph{animated portrait} image $I_A$. 
 We identify that all desired information required for $I_A$ can be extracted from multi-scale perspectives on both the source and target, including: 1) the global head and hair structure information of $I_S$; 2) the identity details on the faces of $I_S$; 3) the global pose of $I_T$; and 4) the detailed expressions of $I_T$.

To this end, we design the Head2Head Aligner as illustrated in Fig. \ref{fig:lpd_framework_fortrain}. 
The Head2Head Aligner consists of a portrait encoder ${E_{por}}$, an identity encoder ${E_{id}}$, a pose encoder ${E_{pose}}$, an expression encoder ${\it{E}_{exp}}$, and a generator $G$.

It is worth noting that during training, the source image and target image are obtained from the same video of a particular person (illustrated in Fig. \ref{fig:lpd_framework_fortrain}), and they are derived from videos of different people during testing. Thus the information disentanglement should be taken into consideration.
 In the training stage, 
we randomly select $K+1$ frames from the same video sequence of an individual person, where $K$ frames are set as the $K$-shot source images $I_S$, while the remaining frame is set as the target image $I_T$.

\noindent\textbf{Source Encoding.} Both the coarse and fine scales are leveraged for information encoding from the source and target. As for the source image, global portrait 
information including the head and hair, is directly extracted by an encoder ${E_{por}}$ to a ${d_{por}}$-dimensional vector ${f_{por}}_k = E_{por}(I_{{S}_{\scriptsize k}})$. 

On the other hand, the identity encoder is adopted from a pretrained state-of-the-art face recognition model \cite{deng2019arc}, which can 
provide more representative identity embedding 
\cite{li2019fs, chen2020simswap, Zhou2021CVPR}.
More specifically, for each source image $I_{{S}_{\scriptsize k}}$, it first undergoes a central cropping transformation $C$, after which $C(I_{{S}_{\scriptsize k}})$ is passed through the identity encoder $E_{id}$ to generate a $d_{id}$-dimensional identity embedding ${f_{id}}_k = {E_{id}(C(I_{{S}_{\scriptsize k}}))}$. 

For randomly fetched $K$ frames of source images ${{I_{{S}_{\scriptsize 1}}},{I_{{S}_{\scriptsize 2}}},...,{I_{{S}_{\scriptsize 
K}}}}$, a total of $K$ portrait vectors ${f_{por}}_k$ and $K$ identity embeddings ${{f_{id}}_k}$ are produced; we acquire ${{\it{\bar{f}_{por}}}}$ and ${{\it{\bar{f}}_{id}}}$ by taking the average of $\{{{f_{por}}_1}, ..., {{f_{por}}_K}\}$ and $\{{{f_{id}}_1}, ..., {{f_{id}}_K}\}$ respectively. 


\noindent\textbf{Target Encoding.}
To obtain both facial expression details and the global head pose, as shown at the bottom-left of Fig. \ref{fig:lpd_framework_fortrain}, we apply the pose encoder ${\it{E}_{pose}}$ and the expression encoder ${\it{E}_{exp}}$ to extract the pose and detailed facial expression information respectively from the target image ${I}_T$. It first undergoes a series of augmentations \cite{Burkov2020lpd} $\it{A}$ before being 
sent into the pose encoder ${\it{E}_{pose}}$. An identity-agnostic $\it{d}_{pose}$-dimensional pose descriptor ${f_{pose}} = {{\it{E}_{pose}} (A({I}_T))}$ is thus encoded. 
Then the augmented image is also processed with central cropping $\it{C}$ mentioned above. The result ${C(A({I}_{T}))}$ is fed into the expression encoder $E_{exp}$ to produce a ${d_{exp}}$-dimensional expression vector $f_{exp}$. 

\noindent\textbf{Animated Portrait Generation.}
The feature descriptors containing information regarding the body, identity, pose and expression are then composed together to produce the \emph{animated portrait}. Specifically, the features ${{\it{\bar{f}_{por}}}}$, ${{\it{\bar{f}}_{id}}}$, ${{\it{{f}}_{pose}}}$ and ${{\it{{f}}_{exp}}}$ are concatenated and processed by a generator via AdaIN\cite{huang2018adain} to produce the desired output $I_A$: 
\begin{align}
    I_A = G({{\it{\bar{f}_{por}}}}, {{\it{\bar{f}_{id}}}}, f_{pose}, f_{exp})
\end{align}
 The loss functions used are basically the reconstruction loss, perceptual loss, adversarial loss and identity loss, which are the same as~\cite{Burkov2020lpd}.
Please refer to supplementary material for more details.

\begin{figure}[t]
\centering
      \includegraphics[width=0.99\linewidth]{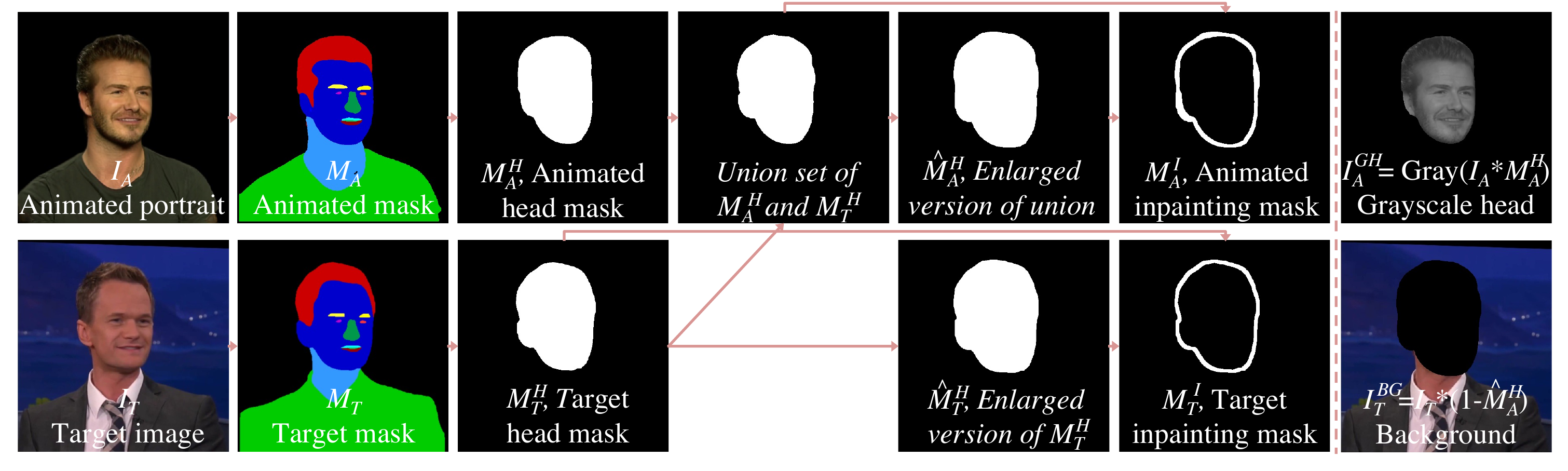}
   \caption{{Illustration of data preprocessing for head2scene blender.}}
   \label{fig:data_preprocess}
   \vspace{-10pt}
\end{figure}

\begin{figure*}
\centering
		\includegraphics[width=0.95\linewidth]{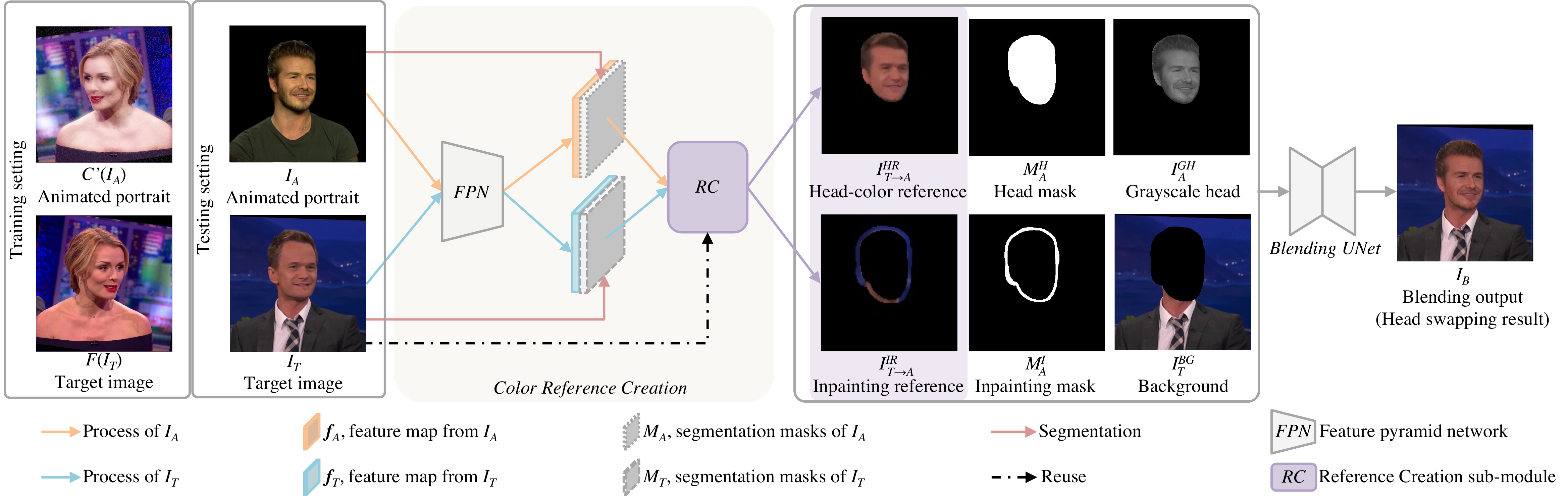}
 	\caption{Illustration of our proposed Head2Scene Blender in the testing setting. The head-color reference $I_{T \rightarrow A}^{HR}$ and inpainting reference $I_{T \rightarrow A}^{IR}$ are first obtained using the Reference Creation sub-module. Then $I_{T \rightarrow A}^{HR}, I_{T \rightarrow A}^{IR}, M_A^H, I_T^{BG}, M_A^I$, and $I_A^{GH}$ are fed into the blending UNet to produce the final result ${I_B}$. In the training phase, the same image is used for $I_A$, $I_T$ and $I_B$.  
	}
	\vspace{-5pt}
	\label{fig:blending_framework-fortest}
\end{figure*}

\subsection{Head2Scene Blender}\label{arch-blending}

With the \emph{animated portrait} image $I_A$ above, our second step is to produce the blending output $I_B$. 
It should be a seamless combination of the head from $I_A$ and the background and body in $I_T$. The challenges are 1) the head shape on $I_A$ could be significantly different from that of $I_T$, thus the blending procedure requires amending the ``outliners'', \textit{i.e.}, masking out the areas of the original head and fill in the gaps led by head shape mismatches;
and 2) the color of the $I_B$ should be consistent with $I_T$.

To this end, we design the Head2Scene Blender, which simultaneously solves both challenges as illustrated in Fig. \ref{fig:blending_framework-fortest}. The key is to build ``\emph{color references}'' which serve as guidance for both skin color transfer and background gap inpainting with a \emph{Semantic-Guided Color Reference Creation} sub-module. Then the final results are generated through a \emph{Blending UNet}. Below we will first introduce the necessary \emph{Data preprocessing}  steps for producing the \emph{color references} and then introduces the sub-modules respectively.

%

\noindent\textbf{Data preprocessing.} The data preprocessing procedure prepares the materials for performing the color transfer and inpainting, which are binary masks that identifies the areas in need and the cropped images. Particularly, 
instead of directly modifying the animated head in $I_A$, we argue that it would be easier to reconfigure the color transfer problem into the problem of re-coloring its grey-scale version of the head $I_A^{GH}$ only. 
Detailedly,
given the \emph{animated portrait} $I_A$, we employ a parsing tool to obtain the segmentation masks $M_A$, which indicates the semantic areas of the portrait image. Then the head mask $M_A^H$ of the target image is obtained by combining the masks belonging to the head area.
Similarly, the segmentation masks $M_T$ and head mask $M_T^H$ of the target image $I_T$ are obtained in the same way. The greyscale head can be computed as $I_A^{GH} = Gray(I_A * M_A^H)$.

Furthermore, the head mask $M_T^H$ is dilated to an enlarged version $\hat{M}_T^H$, then the \emph{target inpainting mask} $M_T^I = \hat{M}_T^H - M_T^H$ which will be further used for creating the \emph{color reference} for inpainting. 
Different from the target inpainting mask above, the \emph{animated inpainting mask} $M_A^I$ is obtained via both $M_A^H$ and $M_T^H$. More especially, the union set of ${M_A^H}$ and ${M_T^H}$ is dilated to an enlarged mask ${\it{\hat{M}}_A^H}$. 
Let the \emph{animated inpainting mask} $M_A^I = {\it{\hat{M}}_A^H} - M_A^H$ denotes the area that needs inpainting when blending $I_A$ with $I_T$. $I^{BG}_T = I_T * (1.0 - \it{\hat{M}}_A^H)$ is the background image without the target head.

%

\noindent\textbf{Semantic-Guided Color Reference Creation.} 
After the data processing steps, we need to re-color the greyscale head  $I_A^{GH}$ and fill the \emph{animated inpainting mask} $M_A^I$.
As stated before, our intuition is to provide a color reference for both of them, namely the \emph{head-color reference} $I_{T \rightarrow A}^{HR}$ and the \emph{inpainting reference} $I_{T \rightarrow A}^{IR}$, respectively. 

Inspired by previous work~\cite{zhang2020cross,zhang2019devc,jiang2020psgan}, we identify that each \emph{color reference} should spatially match the target region and provide pixel-wise color guidance. As the desired color should be consistent with the target image $I_T$, the color of the \emph{references} could be derived by querying the coherent color from $I_T$ through correlation learning. 
This is achieved in the Reference Creation sub-module $RC$.

Due to the lack of paired data, 
the Reference Creation sub-module is still {trained} in a self-driven manner with $I_A$ and $I_T$ sourcing from the \textbf{same} image.
As shown in the left-hand diagram of Fig. \ref{fig:blending_framework-fortest}, 
 in the training phase, the \emph{animated portrait} $I_A$ and target image $I_T$ first undergo a random color augmentation $C'$ and a random horizontal flip augmentation $F$ respectively, which prevents the networks from directly leveraging the pixels from the same position.
The augmented \emph{animated portrait} $C'(I_A)$ and the random flipped \emph{target image} ${F(I_T)}$ are then both sent into a feature pyramid network ${FPN}$ \cite{zhang2020cross} to produce the following semantic representation:
\begin{align}
    {f}_A &= FPN(C'(I_A)). \\
	{f}_T &= FPN(F(I_T)).
    \label{eq:domain_adaptation}
\end{align}


The next step to colorization is to compute the correlations between extracted features on each spatial location, and resample pixel colors from the target image to the \emph{color references}. Previous approaches \cite{zhang2020cross,zhang2019devc,jiang2020psgan} produce a prohibitively large memory footprint when estimating the correspondence due to the fact that the pairwise similarities are computed among all locations of the feature maps, despite their semantic independence. 
As the pixels in the image pairs with different semantic labels make little contribution to the correspondence matrix,
we 
only compute the \textbf{correlations among the same semantic regions} individually. Our practice not only alleviates memory consumption but also avoids mismatched correlations.

\begin{figure}[t]
\centering
      \includegraphics[width=0.95\linewidth]{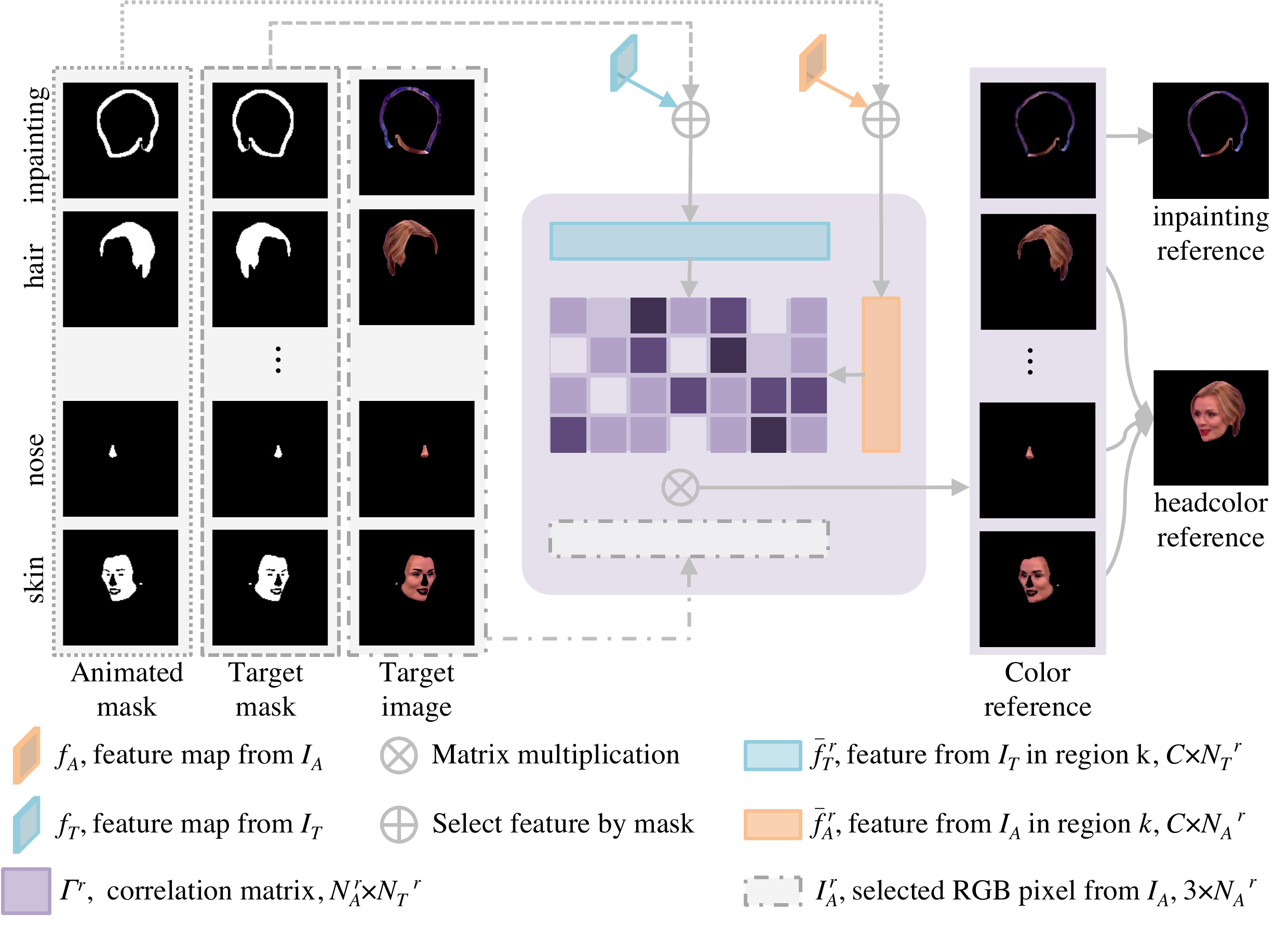}
   \caption{The illustration of our Semantic-Guided Color Reference Creation module, best viewed in color. 
   }
   \label{fig:blending_framework_menorysaving}
\end{figure}

Concretely, as shown in Fig. \ref{fig:blending_framework_menorysaving}, for each \textbf{semantic region} $r$ ${\in}$ $\{ face, hair, eye, nose, lip, tooth, inpainting \}$,
we compute a correlation matrix $\it{\Gamma}^{r} \in \rm{R}^{\it{N}_{A}^{r} \times \it{N}_T^{r}}$, of which each element $\it{\Gamma^r}(u, v)$ is a pairwise feature correlation calculated as:  
\begin{align}
    \it{\Gamma^r}(u, v) = \frac{\bm{\bar{f}}^r_{A}(u)^T\bar{\bm{f}}^r_T(v)}{\Vert{\bar{\bm{f}}^r_{A}(u)} \Vert  \ \Vert{\bar{\bm{f}}^r_T(v)}\Vert}, u \in M_{A}^r, v \in M_T^r,
    \label{eq:correlation_matrix}
\end{align}
where $\bar{{f}}_{A}^r(u)$ and $\bar{{f}}_T^r(v) \in R^{C}$ represent the channel-wise centralized feature of $\it{f}_{A}^r$ and $\it{f}_T^r$ in positions $u$ and $v$ respectively, \ie, $\bar{f}_{A}^r(u) = \it{f}_{A}^r(u) - \text{mean}(f_{A}^r(u))$ and $\bar{f}_T^r(v) = f_T^r(v) - \text{mean}(f_T^r(v))$,  
and the dimensions $\it{N}_{A}^{r}$ and $\it{N}_T^{r}$ are the pixel number of the {semantic region} $\it{r}$ in the animated portrait image and target image respectively.
Compared to the original correlation maps in prior works \cite{zhang2020cross, jiang2020psgan}, our approach reduces the computational complexity from $\it{O}(wh * wh)$ to $\it{O}(\it{N}_{S}^{r} * \it{N}_{R}^{r})$, where $w$ and $h$ denote the spatial size of the original feature map.

As $\it{\Gamma^r}(u, v)$ indicates the similarity between $M_{A}(u)$ and $M_T(v)$, we normalize the $\it{\Gamma^r}(u, v)$ via softmax and compute their weighted contribution of these variables from the target image to the head and inpainting regions of the animated portrait image, as follows:
\begin{equation}
    I_{T \rightarrow A}^r(u) = \sum_{v \in M_T^r} {softmax}_v (\Gamma^r(u,v)/{\tau}) \cdot I_T(v), u \in M_{A}^r.
    \label{eq:warping}
\end{equation}
where  $\tau$ is the temperature coefficient. Finally, as shown in Fig. \ref{fig:blending_framework_menorysaving}, the created inpainting reference $I_{T \rightarrow A}^{IR}$ is derived from $I_T$ and $M_T^I$, while the warped head-color reference $I_{T \rightarrow A}^{HR}$ is obtained from $I_T$ and the combination of $M_T^r$, where $r$ ${\in}$ $\{ face, hair, eye, nose, lip, tooth \}$.
We denote the overall estimation procedure as:
\begin{align}
    I_{T \rightarrow A}^{HR} &= RC(f_A, M_A, f_T, M_T, I_T*M_T^H).  \\
	I_{T \rightarrow A}^{IR} &= RC(f_A, M_A^I, f_T, M_T^I, I_T*M_T^I).
    \label{eq:domain_adaptation}
\end{align}

\noindent\textbf{Blending UNet.} 
After acquiring the created head-color/inpainting reference $I_{T \rightarrow A}^{HR}$/$I_{T \rightarrow A}^{IR}$, the head mask ${M_A^H}$, background ${I_T^{BG}}$, inpainting mask ${M_A^I}$ and gray-scale head ${I_A^{GH}}$ produced by the data processing, we concatenate them all channel-wisely, and pass them to a Blending UNet. Our goal is that:
$1)$ 
the head-color distilled from the target image is transferred to the gray-scale head ${I_A^{GH}}$, where the identity is retained while the color is kept consistent with the remaining skin (such as neck) exposed in the rest of the reference body; $2)$ the missing region masked by the inpainting mask can be estimated. 
Consequently, the overall process of the Blending UNet $B$ can be formulated as follows:
\begin{equation}
    I_B = B(I_{T \rightarrow A}^{HR}, I_{T \rightarrow A}^{IR}, M_A^H, I_T^{BG}, M_A^I, I_A^{GH}).
    \label{eq:blending}
\end{equation}

\noindent\textbf{Cycle Loss}: The training relies on the reconstruction loss, perceptual loss, adversarial loss. Additionally,
in order to guarantee that the warped headcolor/inpainting exemplars could learn a meaningful correspondence matrix, we introduce the cycle consistent loss.
\begin{equation}
    L_c = \lambda_{c}\left\|{I_{T \rightarrow A \rightarrow T}-I_T}\right\|_1,
\end{equation}
where $I_{T \rightarrow A \rightarrow T}$ is the color reference after cycled, and $I_{T \rightarrow A \rightarrow T}^k(u) = \sum_{v \in M_A^k} {softmax}_v (\Gamma^k(u,v)/{\tau}) \cdot I_{T \rightarrow A}(v), u \in M_T^k$. Besides, the additional target image $I_T'$ coming from different image compared to $I_A$ is also utilized to ensure the meaningful of warped exemplar:
\begin{equation}
    L_{c'} = \lambda_{c}\left\|{I_{T' \rightarrow A \rightarrow T'}-I_T}\right\|_1.
\end{equation}
Please refer to the supplementary materials for more details.


%% file: main.bbl
\begin{thebibliography}{10}\itemsep=-1pt

\bibitem{brock2018large}
Brock Andrew, Donahue Jeff, and Simonyan Karen.
\newblock Large scale {GAN} training for high fidelity natural image synthesis.
\newblock {\em arXiv preprint arXiv:1809.11096}, 2018.

\bibitem{Bao2018ipgan}
Jianmin Bao, Dong Chen, Fang Wen, Houqiang Li, and Gang Hua.
\newblock Towards open-set identity preserving face synthesis.
\newblock In {\em IEEE Conf. Comput. Vis. Pattern Recog.}, 2018.

\bibitem{Change2018pairedcyclegan}
Huiwen Chang, Jingwan Lu, Fisher Yu, and Adam Finkelstein.
\newblock Pairedcyclegan: Asymmetric style transfer for applying and removing
  makeup.
\newblock {\em cvpr}, 2018.

\bibitem{chen2020simswap}
Renwang Chen, Xuanhong Chen, and Yanhao. Ge.
\newblock Simswap: An efficient framework for high fidelity face swapping.
\newblock In {\em ACM Int. Conf. Multimedia}, 2020.

\bibitem{Chen_2020_CVPR}
Zhuo Chen, Chaoyue Wang, Bo Yuan, and Dacheng Tao.
\newblock Puppeteergan: Arbitrary portrait animation with semantic-aware
  appearance transformation.
\newblock In {\em Proceedings of the IEEE/CVF Conference on Computer Vision and
  Pattern Recognition (CVPR)}, 2020.

\bibitem{deng2019arc}
Jiankang Deng, Jia Guo, Niannan Xue, and Stefanos. Zafeiriou.
\newblock Arcface: Additive angular margin loss for deep face recognition.
\newblock In {\em IEEE Conf. Comput. Vis. Pattern Recog.}, 2019.

\bibitem{Burkov2020lpd}
Burkov Egor, Pasechnik Igor, Grigorev Artur, and Lempitsky Victor.
\newblock Neural head reenactment with latent pose descriptors.
\newblock In {\em IEEE Conf. Comput. Vis. Pattern Recog.}, 2020.

\bibitem{gafni20dnrf}
Guy Gafni, Justus Thies, Michael Zollhofer, and Matthias Nie{\ss}ner.
\newblock Dynamic neural radiance fields for monocular 4d facial avatar
  reconstruction.
\newblock {\em arXiv preprint arXiv:2012.03065}, 2020.

\bibitem{goodfellow2014generative}
Ian Goodfellow, Jean Pouget-Abadie, Mehdi Mirza, Bing Xu, David Warde-Farley,
  Sherjil Ozair, Aaron Courville, and Yoshua Bengio.
\newblock Generative adversarial nets.
\newblock {\em Advances in neural information processing systems}, 27, 2014.

\bibitem{gu2019ladn}
Qiao Gu, Guanzhi Wang, Tik~Chiu Mang, YuWing Tai, and ChiKeung Tang.
\newblock Ladn: Local adversarial disentangling network for facial makeup and
  de-makeup.
\newblock {\em iccv}, 2019.

\bibitem{deng2021scgan}
Han, Chu Deng, Hongmin Han, Guoqiang cai, Shengfeng Han, and He.
\newblock Spatially-invariant style-codes controlled makeup transfer.
\newblock In {\em CVPR}, 2021.

\bibitem{zhang2018self}
Zhang Han, Goodfellow Ian, Metaxas Dimitris, and Odena Augustus.
\newblock Self-attention generative adversarial networks.
\newblock {\em arXiv preprint arXiv:1805.08318}, 2018.

\bibitem{huang2018adain}
Xun Huang and Serge. Belongie.
\newblock Arbitrary style transfer in real-time with adaptive instance
  normalization.
\newblock In {\em IEEE Conf. Conference of the International Speech
  Communication Association.}, 2017.

\bibitem{chang2018pairedcyclegan}
Huiwen, Jingwan Chang, Fisher Lu, Adam Yu, and Finkelstein.
\newblock Pairedcyclegan: Asymmetric style transfer for applying and removing
  makeup.
\newblock In {\em CVPR}, 2018.

\bibitem{chen2019beautyglow}
Hungjen, Kaming Chen, Szuyu Hui, Liwu Wang, Honghan Tsao, Wenhuang Shuai, and
  Cheng.
\newblock Beautyglow: On-demand makeup transfer framework with reversible
  generative network.
\newblock In {\em CVPR}, 2019.

\bibitem{JiCVPR2021}
Xinya Ji, Hang Zhou, Kaisiyuan Wang, Wayne Wu, Chan~Change Loy, Xun Cao, and
  Feng Xu.
\newblock Audio-driven emotional video portraits.
\newblock In {\em The IEEE Conference on Computer Vision and Pattern
  Recognition (CVPR)}, 2021.

\bibitem{jiang2020psgan}
Wentao Jiang, Si Liu, Chen Gao, Jie Cao, Ran He, Jiashi Feng, and Shuicheng
  Yan.
\newblock X2face: A network for controlling face generation by using images,
  audio, and pose codes.
\newblock {\em cvpr}, 2020.

\bibitem{chung2018vox2}
Son~Chung Joon, Nagrani Arsha, and Zisserman. Andrew.
\newblock Voxceleb2: Deepspeaker recognition.
\newblock In {\em IEEE Conf. Conference of the International Speech
  Communication Association.}, 2018.

\bibitem{kim2018dvp}
Hyeongwoo Kim, Pablo Garrido, Ayush Tewari, Weipeng Xu, Justus Thies, Matthias
  Niessner, Patrick P{\'e}rez, Christian Richardt, Michael Zollh{\"o}fer, and
  Christian Theobalt.
\newblock Deep video portraits.
\newblock {\em SIGGRAPH}, 2018.

\bibitem{kingma2014adam}
Diederik~P Kingma and Jimmy Ba.
\newblock Adam: A method for stochastic optimization.
\newblock {\em arXiv preprint arXiv:1412.6980}, 2014.

\bibitem{li2019fs}
Lingzhi Li, Jianmin Bao, Hao Yang, Dong Chen, and Fang. Wen.
\newblock Faceshifter: Towards high fidelity and occlusion aware face swapping.
\newblock In {\em IEEE Conf. Comput. Vis. Pattern Recog.}, 2019.

\bibitem{Li2019faceshifter}
Lingzhi Li, Jianmin Bao, Hao Yang, Dong Chen, and Fang Wen.
\newblock Faceshifter: Towards high fidelity and occlusion aware face swapping.
\newblock In {\em IEEE Conf. Comput. Vis. Pattern Recog.}, 2019.

\bibitem{liang2022expressive}
Borong Liang, Yan Pan, Zhizhi Guo, Hang Zhou, Zhibin Hong, Xiaoguang Han, Junyu
  Han, Jingtuo Liu, Errui Ding, and Jingdong Wang.
\newblock Expressive talking head generation with granular audio-visual
  control.
\newblock In {\em Proceedings of the IEEE/CVF Conference on Computer Vision and
  Pattern Recognition (CVPR)}, 2022.

\bibitem{lu2021live}
Yuanxun Lu, Jinxiang Chai, and Xun Cao.
\newblock Live speech portraits: real-time photorealistic talking-head
  animation.
\newblock {\em ACM Transactions on Graphics (TOG)}, 40(6):1--17, 2021.

\bibitem{Nirkin2019fsgan}
Yuval Nirkin, Yosi Keller, and Tal Hassner.
\newblock Fsgan: Subject agnostic face swapping and reenactment.
\newblock {\em iccv}, 2019.

\bibitem{nirkin2018face}
Yuval Nirkin, Iacopo Masi, Anh~Tran Tuan, Tal Hassner, and Gerard Medioni.
\newblock On face segmentation, face swapping, and face perception.
\newblock In {\em 2018 13th IEEE International Conference on Automatic Face \&
  Gesture Recognition (FG 2018)}, pages 98--105. IEEE, 2018.

\bibitem{Perov2020deepfacelab}
Ivan Perov, Daiheng Gao, Nikolay Chervoniy, Kunlin Liu, Sugasa Marangonda,
  Chris Um, Mr Dpfks, Carl~Shift Facenheim, Luis RP, Jian Jiang, et~al.
\newblock Deepfacelab: Integrated, flexible and extensible face-swapping
  framework.
\newblock {\em arXiv preprint arXiv:2005.05535}, 2020.

\bibitem{Sandler18}
Mark Sandler, Andrew Howard, Menglong Zhu, Andrey Zhmoginov, and Liang-Chieh
  Chen.
\newblock Mobilenetv2: Inverted residuals and linear bottlenecks.
\newblock In {\em IEEE Conf. Comput. Vis. Pattern Recog.}, June 2018.

\bibitem{sarkar2021styleposegan}
Kripasindhu Sarkar, Vladislav Golyanik, Lingjie Liu, and Christian Theobalt.
\newblock Style and pose control for image synthesis of humans from a single
  monocular view.
\newblock {\em arXiv preprint arXiv:2102.11263}, 2021.

\bibitem{liu2021psgan++}
Si, Wentao Liu, Chen Jiang, Ran Gao, Jiashi He, Bo Feng, Shuicheng Li, and Yan.
\newblock Psgan++: Robust detail-preserving makeup transfer and removal.
\newblock In {\em PAMI}, 2021.

\bibitem{siarohin2019animating}
Aliaksandr Siarohin, St{\'e}phane Lathuili{\`e}re, Sergey Tulyakov, Elisa
  Ricci, and Nicu Sebe.
\newblock Animating arbitrary objects via deep motion transfer.
\newblock {\em CVPR}, 2019.

\bibitem{Siarohin2019fomm}
Aliaksandr Siarohin, St{\'e}phane Lathuili{\`e}re, Sergey Tulyakov, Elisa
  Ricci, and Nicu Sebe.
\newblock First order motion model for image animation.
\newblock {\em neurips}, 2019.

\bibitem{siarohin2021motion}
Aliaksandr Siarohin, Oliver Woodford, Jian Ren, Menglei Chai, and Sergey
  Tulyakov.
\newblock Motion representations for articulated animation.
\newblock In {\em CVPR}, 2021.

\bibitem{karras2020stylegan2}
Tero, Samuli Karras, Miika Laine, Janne Aittala, Jaakko Hellsten, Timo
  Lehtinen, and Aila.
\newblock Analyzing and improving the image quality of stylegan.
\newblock In {\em CVPR}, 2020.

\bibitem{thies19dnr}
Justus Thies, Michael Zollh{\"o}fer, and Matthias Nie{\ss}ner.
\newblock Deferred neural rendering: Image synthesis using neural textures.
\newblock {\em ACM Trans. Graph.}, 2019.

\bibitem{li2018beautygan}
Tingting, Ruihe Li, Chao Qian, Si Dong, Qiong Liu, Wenwu Yan, Liang Zhu, and
  Lin.
\newblock Beautygan: Instance-level facial makeup transfer with deep generative
  adversarial network.
\newblock In {\em ACMMM}, 2018.

\bibitem{zhou2004ssim}
Zhou Wang, Alan~C. Bovik, Hamid~R. Sheikh, and Eero~P. Simoncelli.
\newblock Image quality assessment: from error visibility to structural
  similarity.
\newblock In {\em IEEE transactions on image processing}, 2004.

\bibitem{Xie17}
Saining Xie, Ross Girshick, Piotr Dollar, Zhuowen Tu, and Kaiming He.
\newblock Aggregated residual transformations for deep neural networks.
\newblock In {\em IEEE Conf. Comput. Vis. Pattern Recog.}, July 2017.

\bibitem{xu2022mobilefaceswap}
Zhiliang Xu, Zhibin Hong, Changxing Ding, Zhen Zhu, Junyu Han, Jingtuo Liu, and
  Errui Ding.
\newblock Mobilefaceswap: A lightweight framework for video face swapping.
\newblock {\em AAAI}, 2022.

\bibitem{xu2021facecontroller}
Zhiliang Xu, Xiyu Yu, Zhibin Hong, Zhen Zhu, Junyu Han, Jingtuo Liu, Errui
  Ding, and Xiang Bai.
\newblock Facecontroller: Controllable attribute editing for face in the wild.
\newblock {\em AAAI}, 2021.

\bibitem{Zakharov20fast}
Egor Zakharov, Aleksei Ivakhnenko, Aliaksandra Shysheya, and Victor Lempitsky.
\newblock Fast bi-layer neural synthesis of one-shot realistic head avatars.
\newblock In {\em Eur. Conf. Comput. Vis.}, August 2020.

\bibitem{Zakharov19fsth}
Egor Zakharov, Aliaksandra Shysheya, Egor Burkov, and Victor Lempitsky.
\newblock Few-shot adversarial learning of realistic neural talking head
  models.
\newblock In {\em Int. Conf. Comput. Vis.}, October 2019.

\bibitem{zhang2019devc}
Bo Zhang, Mingming He, Jing Liao, V.~Sander Pedro, Lu Yuan, Amine Bermak, and
  Dong Chen.
\newblock Deep exemplar-based video colorization.
\newblock In {\em CVPR}, pages 8052--8061, 2019.

\bibitem{zhang2020freenet}
Jiangning Zhang, Xianfang Zeng, Mengmeng Wang, Yusu Pan, Liang Liu, Yong Liu,
  Yu Ding, and Changjie Fan.
\newblock Freenet: Multi-identity face reenactment.
\newblock In {\em Proceedings of the IEEE/CVF Conference on Computer Vision and
  Pattern Recognition (CVPR)}, 2020.

\bibitem{zhang2020cross}
Pan Zhang, Bo Zhang, Dong Chen, Lu Yuan, and Fang Wen.
\newblock Cross-domain correspondence learning for exemplar-based image
  translation.
\newblock In {\em CVPR}, pages 5143--5153, 2020.

\bibitem{zhang2018unreasonable}
Richard Zhang, Phillip Isola, Alexei~A Efros, Eli Shechtman, and Oliver Wang.
\newblock The unreasonable effectiveness of deep features as a perceptual
  metric.
\newblock In {\em IEEE Conf. Comput. Vis. Pattern Recog.}, pages 586--595,
  2018.

\bibitem{zhang2021flow}
Zhimeng Zhang, Lincheng Li, Yu Ding, and Changjie Fan.
\newblock Flow-guided one-shot talking face generation with a high-resolution
  audio-visual dataset.
\newblock In {\em Proceedings of the IEEE/CVF Conference on Computer Vision and
  Pattern Recognition (CVPR)}, 2021.

\bibitem{zhou2019talking}
Hang Zhou, Yu Liu, Ziwei Liu, Ping Luo, and Xiaogang Wang.
\newblock Talking face generation by adversarially disentangled audio-visual
  representation.
\newblock In {\em Proceedings of the AAAI Conference on Artificial Intelligence
  (AAAI)}, 2019.

\bibitem{Zhou2021CVPR}
Hang Zhou, Yasheng Sun, Wayne Wu, Chen~Change Loy, Xiaogang Wang, and Ziwei
  Liu.
\newblock Pose-controllable talking face generation by implicitly modularized
  audio-visual representation.
\newblock In {\em Proceedings of the IEEE Conference on Computer Vision and
  Pattern Recognition (CVPR)}, 2021.

\bibitem{zhou2020loa}
Tong Zhou, Changxing Ding, Shaowen Lin, Xinchao Wang, and Dacheng Tao.
\newblock Learning oracle attention for high-fidelity face completion.
\newblock {\em CVPR}, 2020.

\bibitem{zhou2021transfill}
Yuqian Zhou, Connelly Barnes, Eli Shechtman, and Sohrab Amirghodsi.
\newblock Transfill: Reference-guided image inpainting by merging multiple
  color and spatial transformations.
\newblock {\em CVPR}, pages 2266--2276, 2021.

\bibitem{Zhu_2021_CVPR}
Yuhao Zhu, Qi Li, Jian Wang, Cheng-Zhong Xu, and Zhenan Sun.
\newblock One shot face swapping on megapixels.
\newblock In {\em Proceedings of the IEEE/CVF Conference on Computer Vision and
  Pattern Recognition (CVPR)}, pages 4834--4844, 2021.

\end{thebibliography}
